\documentclass[notitlepage]{report}
\usepackage{amsmath,amsfonts,amsthm,amssymb}
\usepackage{authblk}
\usepackage[T1]{fontenc}
\usepackage{graphicx}
\usepackage[utf8]{inputenc}
\usepackage[framemethod=tikz]{mdframed}
\usepackage{natbib}
\usepackage{subcaption}
\usepackage{tikz}
\usepackage{xcolor}
\usepackage{epigraph}
\usepackage{float}

\usepackage{hyperref}

\definecolor{blue}{RGB}{38,139,210}
\definecolor{cyan}{RGB}{42,161,152}
\definecolor{violet}{RGB}{108,113,196}
\definecolor{red}{RGB}{220,50,47}
\definecolor{base01}{RGB}{88,110,117}
\definecolor{base02}{RGB}{7,54,66}
\definecolor{base03}{RGB}{0,43,54}

\usetikzlibrary{calc,shapes,positioning}

\newtheorem{relationship}{Relationship}

\surroundwithmdframed[
    topline=false,
    bottomline=false,
    middlelinewidth=0.5pt,
    linecolor=base01,
    roundcorner=5pt,
    innertopmargin=0pt,
    leftmargin=15pt,
    rightmargin=15pt,
    nobreak=true,
]{relationship}

\let\originalepigraph\epigraph
\renewcommand\epigraph[2]{\originalepigraph{\textit{#1}}{\textsc{#2}}}

\makeatletter
\let\@fnsymbol\@arabic
\makeatother

\title{A guide to convolution arithmetic for deep learning}
\author[$\bigstar$]{Vincent Dumoulin\thanks{dumouliv@iro.umontreal.ca}}
\author[$\bigstar\dagger$]{Francesco Visin\thanks{francesco.visin@polimi.it}}
\affil[$\bigstar$]{MILA, Universit\'{e} de Montr\'{e}al}
\affil[$\dagger$]{AIRLab, Politecnico di Milano}
\date{\today}

\begin{document}

\maketitle
\thispagestyle{empty}
\clearpage

\setlength{\epigraphwidth}{0.4\textwidth}
\epigraph{All models are wrong, but some are useful.}{George E. P. Box}
\clearpage

\renewcommand{\abstractname}{Acknowledgements}
\begin{abstract}
    The authors of this guide would like to thank David Warde-Farley, Guillaume
    Alain and Caglar Gulcehre for their valuable feedback. We are likewise
    grateful to all those who helped improve this tutorial with helpful
    comments, constructive criticisms and code contributions. Keep them coming!

    Special thanks to Ethan Schoonover, creator of the Solarized color
    scheme,\footnote{\url{http://ethanschoonover.com/solarized}} whose colors
    were used for the figures.
\end{abstract}

\renewcommand{\abstractname}{Feedback}
\begin{abstract}
    Your feedback is welcomed! We did our best to be as precise, informative and
    up to the point as possible, but should there be anything you feel might be
    an error or could be rephrased to be more precise or comprehensible, please
    don't refrain from contacting us. Likewise, drop us a line if you think
    there is something that might fit this technical report and you would like
    us to discuss -- we will make our best effort to update this document.
\end{abstract}

\renewcommand{\abstractname}{Source code and animations}
\begin{abstract}
    The code used to generate this guide along with its figures is available on
    GitHub.\footnote{\url{https://github.com/vdumoulin/conv_arithmetic}} There
    the reader can also find an animated version of the figures.
\end{abstract}

\tableofcontents

\chapter{Introduction}

Deep convolutional neural networks (CNNs) have been at the heart of spectacular
advances in deep learning. Although CNNs have been used as early as the nineties
to solve character recognition tasks \citep{le1997reading}, their current
widespread application is due to much more recent work, when a deep CNN was used
to beat state-of-the-art in the ImageNet image classification challenge
\citep{krizhevsky2012imagenet}.

Convolutional neural networks therefore constitute a very useful tool for
machine learning practitioners. However, learning to use CNNs for the first time
is generally an intimidating experience. A convolutional layer's output shape is
affected by the shape of its input as well as the choice of kernel shape, zero
padding and strides, and the relationship between these properties is not
trivial to infer. This contrasts with fully-connected layers, whose output size
is independent of the input size. Additionally, CNNs also usually feature a {\em
pooling\/} stage, adding yet another level of complexity with respect to
fully-connected networks.  Finally, so-called transposed convolutional layers
(also known as fractionally strided convolutional layers) have been employed in
more and more work as of late \citep{zeiler2011adaptive,zeiler2014visualizing,
long2015fully,radford2015unsupervised,visin15,im2016generating}, and their
relationship with convolutional layers has been explained with various degrees
of clarity.

This guide's objective is twofold:

\begin{enumerate}
    \item Explain the relationship between convolutional layers and transposed
        convolutional layers.
    \item Provide an intuitive understanding of the relationship between input
        shape, kernel shape, zero padding, strides and output shape in
        convolutional, pooling and transposed convolutional layers.
\end{enumerate}

In order to remain broadly applicable, the results shown in this guide are
independent of implementation details and apply to all commonly used machine
learning frameworks, such as Theano
\citep{bergstra2010theano,bastien2012theano}, Torch \citep{collobert2011torch7},
Tensorflow \citep{abaditensorflow} and Caffe \citep{jia2014caffe}.

This chapter briefly reviews the main building blocks of CNNs, namely discrete
convolutions and pooling. For an in-depth treatment of the subject, see Chapter
9 of the Deep Learning textbook \citep{Goodfellow-et-al-2016-Book}.

\section{Discrete convolutions}

The bread and butter of neural networks is \emph{affine transformations}: a
vector is received as input and is multiplied with a matrix to produce an
output (to which a bias vector is usually added before passing the result
through a nonlinearity). This is applicable to any type of input, be it an
image, a sound clip or an unordered collection of features: whatever their
dimensionality, their representation can always be flattened into a vector
before the transformation.

Images, sound clips and many other similar kinds of data have an intrinsic
structure. More formally, they share these important properties:

\begin{itemize}
    \item They are stored as multi-dimensional arrays.
    \item They feature one or more axes for which ordering matters (e.g., width
        and height axes for an image, time axis for a sound clip).
    \item One axis, called the channel axis, is used to access different views
        of the data (e.g., the red, green and blue channels of a color image, or
        the left and right channels of a stereo audio track).
\end{itemize}

These properties are not exploited when an affine transformation is applied; in
fact, all the axes are treated in the same way and the topological information
is not taken into account. Still, taking advantage of the implicit structure of
the data may prove very handy in solving some tasks, like computer vision and
speech recognition, and in these cases it would be best to preserve it. This is
where discrete convolutions come into play.

A discrete convolution is a linear transformation that preserves this notion of
ordering. It is sparse (only a few input units contribute to a given output
unit) and reuses parameters (the same weights are applied to multiple locations
in the input).

\begin{figure}[p]
    \centering
    \includegraphics[width=0.32\textwidth]{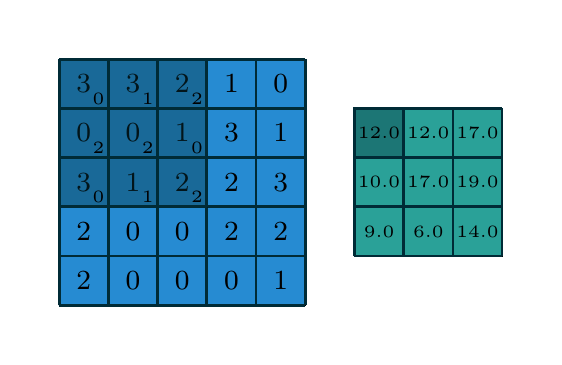}
    \includegraphics[width=0.32\textwidth]{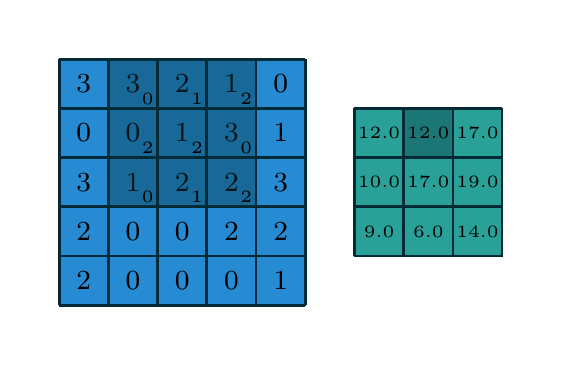}
    \includegraphics[width=0.32\textwidth]{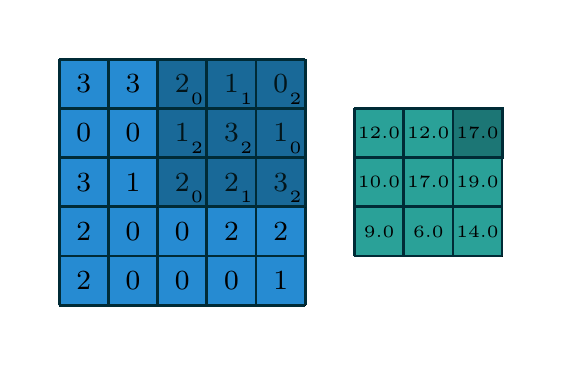}
    \includegraphics[width=0.32\textwidth]{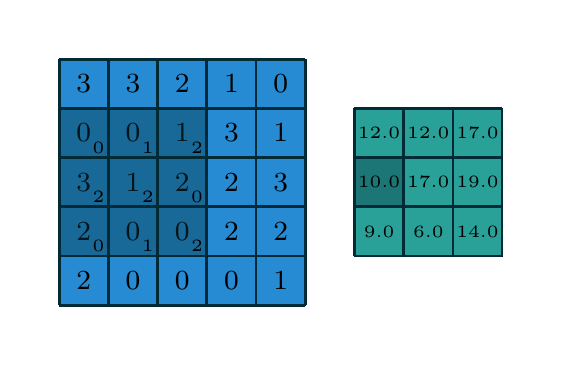}
    \includegraphics[width=0.32\textwidth]{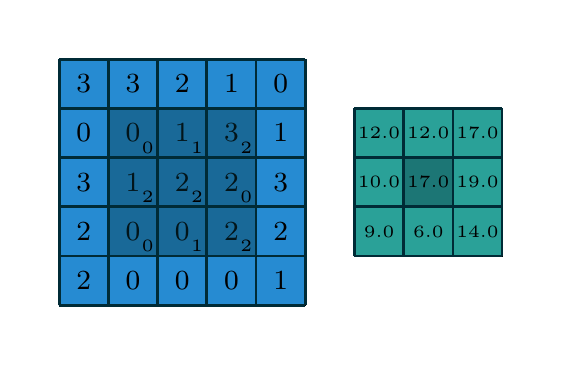}
    \includegraphics[width=0.32\textwidth]{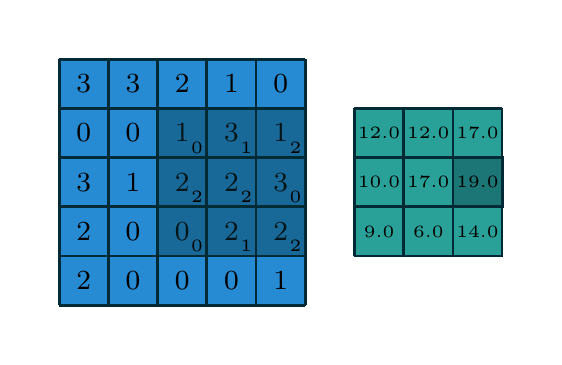}
    \includegraphics[width=0.32\textwidth]{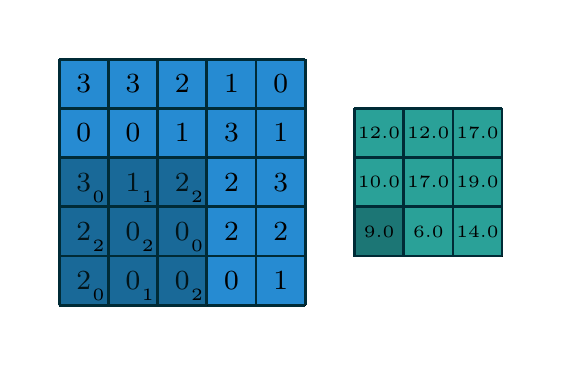}
    \includegraphics[width=0.32\textwidth]{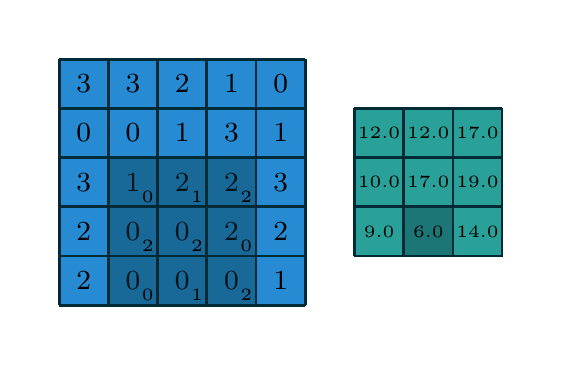}
    \includegraphics[width=0.32\textwidth]{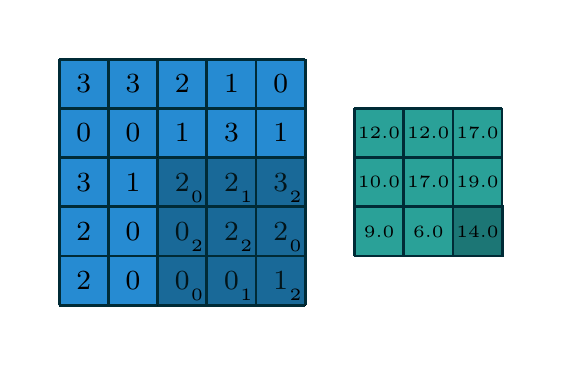}
    \caption{\label{fig:numerical_no_padding_no_strides} Computing the output
        values of a discrete convolution.}
\end{figure}

\autoref{fig:numerical_no_padding_no_strides} provides an example of a discrete
convolution. The light blue grid is called the {\em input feature map}. To keep
the drawing simple, a single input feature map is represented, but it is not
uncommon to have multiple feature maps stacked one onto another.\footnote{%
    An example of this is what was referred to earlier as {\em channels\/} for
    images and sound clips.}
A {\em kernel\/} (shaded area) of value

\begin{figure}[H]
    \centering
    \begin{tikzpicture}[scale=.4,every node/.style={minimum size=1cm}, on grid]
            \draw[fill=base02,opacity=0.4] (0,0) rectangle (3,3);
            \draw[draw=base03,thick] (0,0) grid (3,3);
            \node (00) at (0.5,2.5) {\tiny 0};
            \node (01) at (1.5,2.5) {\tiny 1};
            \node (02) at (2.5,2.5) {\tiny 2};
            \node (10) at (0.5,1.5) {\tiny 2};
            \node (11) at (1.5,1.5) {\tiny 2};
            \node (12) at (2.5,1.5) {\tiny 0};
            \node (20) at (0.5,0.5) {\tiny 0};
            \node (21) at (1.5,0.5) {\tiny 1};
            \node (22) at (2.5,0.5) {\tiny 2};
    \end{tikzpicture}
\end{figure}

\begin{figure}[p]
    \centering
    \includegraphics[width=0.32\textwidth]{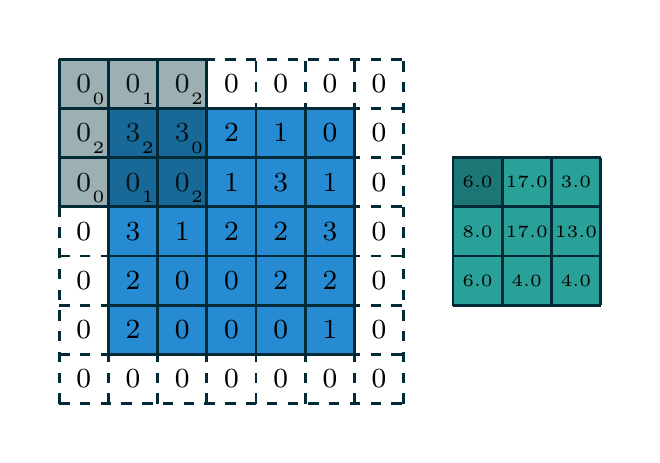}
    \includegraphics[width=0.32\textwidth]{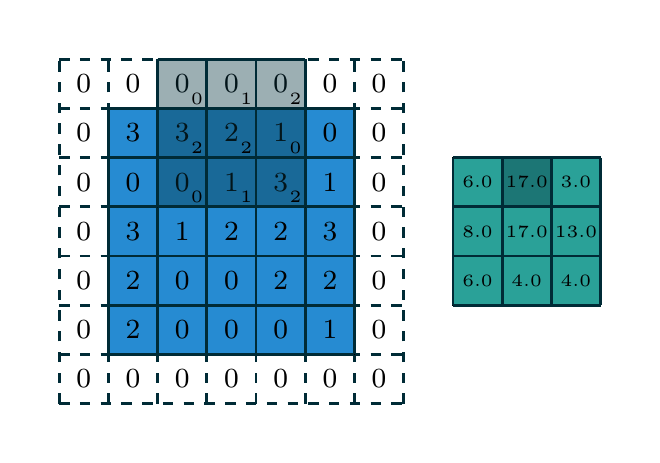}
    \includegraphics[width=0.32\textwidth]{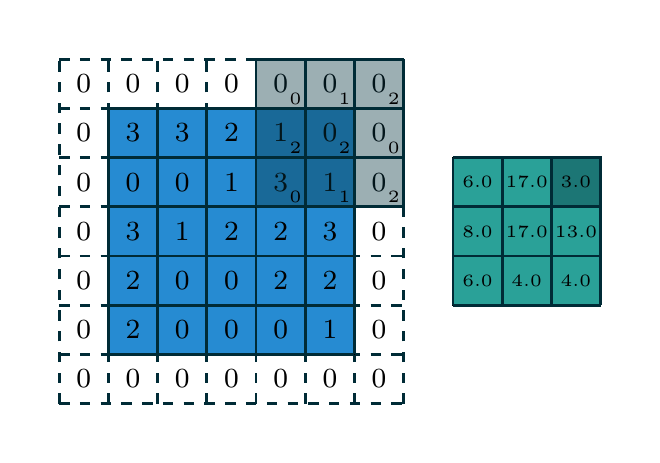}
    \includegraphics[width=0.32\textwidth]{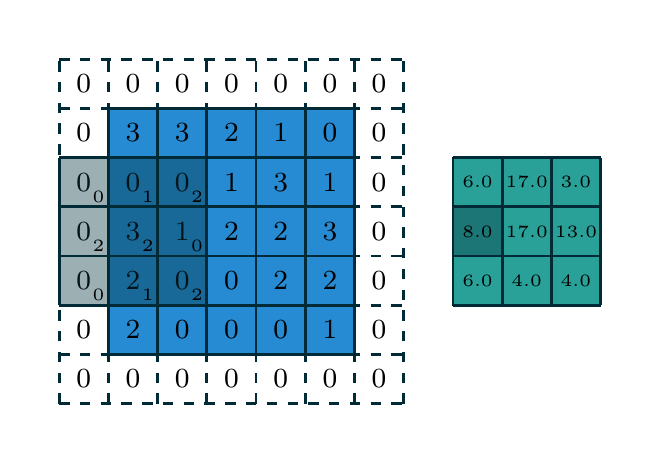}
    \includegraphics[width=0.32\textwidth]{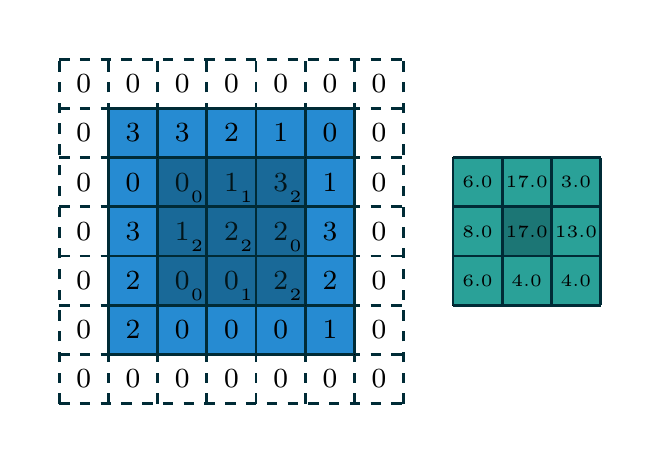}
    \includegraphics[width=0.32\textwidth]{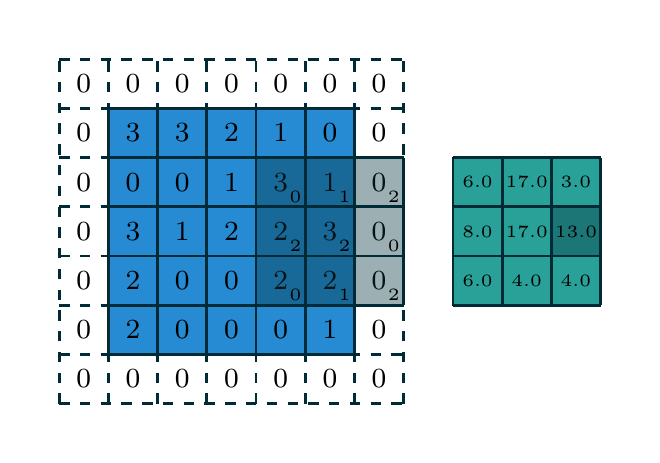}
    \includegraphics[width=0.32\textwidth]{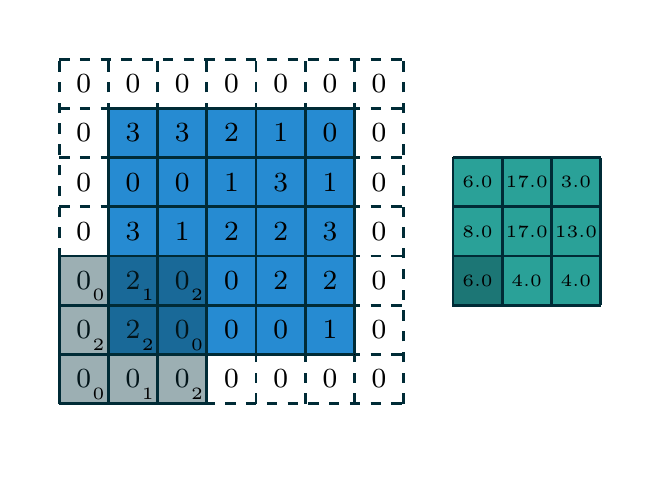}
    \includegraphics[width=0.32\textwidth]{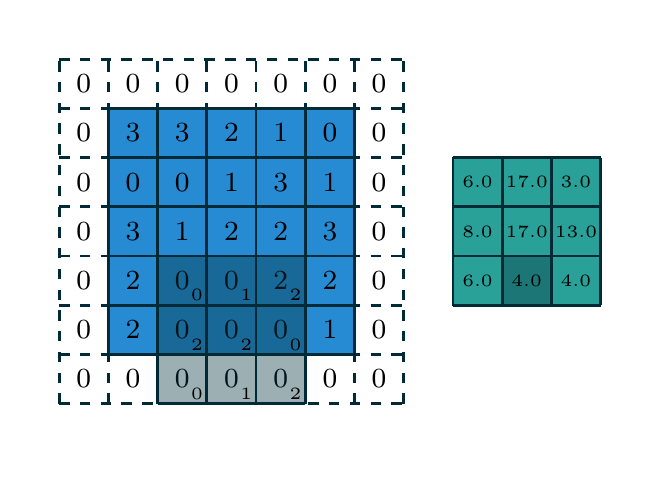}
    \includegraphics[width=0.32\textwidth]{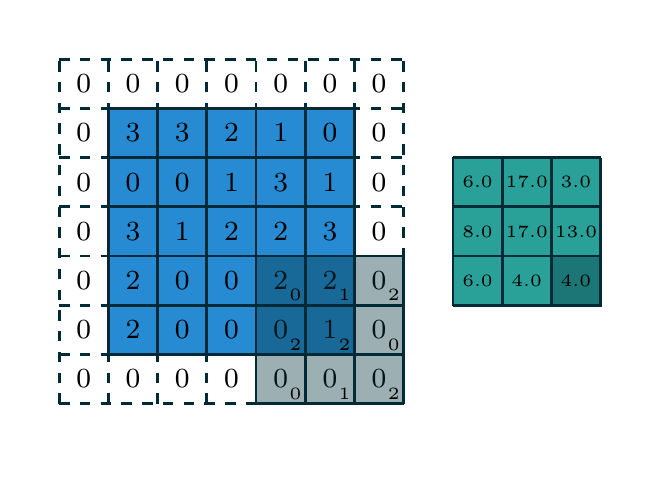}
    \caption{\label{fig:numerical_padding_strides} Computing the output values
        of a discrete convolution for $N = 2$, $i_1 = i_2 = 5$, $k_1 = k_2 = 3$,
        $s_1 = s_2 = 2$, and $p_1 = p_2 = 1$.}
\end{figure}

\noindent slides across the input feature map. At each location, the product
between each element of the kernel and the input element it overlaps is computed
and the results are summed up to obtain the output in the current location. The
procedure can be repeated using different kernels to form as many output feature
maps as desired (\autoref{fig:full_picture}). The final outputs of this procedure
are called {\em output feature maps}.\footnote{%
    While there is a distinction between convolution and cross-correlation from
    a signal processing perspective, the two become interchangeable when the
    kernel is learned. For the sake of simplicity and to stay consistent with
    most of the machine learning literature, the term {\em convolution\/}
    will be used in this guide.}
If there are multiple input feature maps, the kernel will have to be
3-dimensional -- or, equivalently each one of the feature maps will be
convolved with a distinct kernel -- and the resulting feature maps will
be summed up elementwise to produce the output feature map.

The convolution depicted in \autoref{fig:numerical_no_padding_no_strides} is an
instance of a 2-D convolution, but it can be generalized to N-D convolutions.
For instance, in a 3-D convolution, the kernel would be a {\em cuboid\/} and
would slide across the height, width and depth of the input feature map.

The collection of kernels defining a discrete convolution has a shape
corresponding to some permutation of $(n, m, k_1, \ldots, k_N)$, where

\begin{equation*}
\begin{split}
    n &\equiv \text{number of output feature maps},\\
    m &\equiv \text{number of input feature maps},\\
    k_j &\equiv \text{kernel size along axis $j$}.
\end{split}
\end{equation*}

The following properties affect the output size $o_j$ of a convolutional layer
along axis $j$:

\begin{itemize}
    \item $i_j$: input size along axis $j$,
    \item $k_j$: kernel size along axis $j$,
    \item $s_j$: stride (distance between two consecutive positions of the
        kernel) along axis $j$,
    \item $p_j$: zero padding (number of zeros concatenated at the beginning and
        at the end of an axis) along axis $j$.
\end{itemize}

\noindent For instance, \autoref{fig:numerical_padding_strides} shows a $3
\times 3$ kernel applied to a $5 \times 5$ input padded with a $1 \times 1$
border of zeros using $2 \times 2$ strides.

\begin{figure}[p]
    \centering
    \begin{tikzpicture}[scale=.35,every node/.style={minimum size=1cm}, on grid]
        \begin{scope}[xshift=0,yshift=0cm]
            \begin{scope}[xshift=0cm,yshift=0cm]
                \draw[draw=base03,fill=blue,thick] (0,0) grid (5,5) rectangle (0,0);
                \draw[fill=base02, opacity=0.4] (0,2) rectangle (3,5);
            \end{scope}
            \begin{scope}[xshift=7cm,yshift=1.5cm]
                \draw[draw=base03,fill=cyan,thick] (0,0) grid (2,2) rectangle (0,0);
            \end{scope}
        \end{scope}
        \draw[draw=base03, ->, thick] (2.6,3.5) to  (4.5,3.5);
        \draw[draw=base03, ->, thick] (1.5,2.4) to (1.5,0.5);
        \draw[draw=base03, ->, thick] (5.25, 2.5) to (6.75, 2.5);
        \begin{scope}[xshift=12cm,yshift=0cm]
            \begin{scope}[xshift=0cm,yshift=0cm]
                \draw[draw=base03,fill=blue,thick] (0,0) grid (5,5) rectangle (0,0);
                \draw[fill=base02, opacity=0.4] (0,2) rectangle (3,5);
            \end{scope}
            \begin{scope}[xshift=7cm,yshift=1cm]
                \draw[draw=base03,fill=cyan,thick] (0,0) grid (3,3) rectangle (0,0);
                \draw[draw=base03] (1,0) -- (2,1) -- (2,0) -- (1,1);
                \draw[draw=base03] (0,1) -- (1,2) -- (1,1) -- (0,2);
                \draw[draw=base03] (1,1) -- (2,2) -- (2,1) -- (1,2);
                \draw[draw=base03] (2,1) -- (3,2) -- (3,1) -- (2,2);
                \draw[draw=base03] (1,2) -- (2,3) -- (2,2) -- (1,3);
            \end{scope}
            \begin{scope}[xshift=12cm,yshift=1.5cm]
                \draw[draw=base03,fill=cyan,thick] (0,0) grid (2,2) rectangle (0,0);
            \end{scope}
        \end{scope}
        \draw[draw=base03, ->, thick] (14.6,3.5) to  (15.5,3.5);
        \draw[draw=base03, ->, thick] (15.6,3.5) to  (16.5,3.5);
        \draw[draw=base03, ->, thick] (13.5,2.4) to (13.5,1.5);
        \draw[draw=base03, ->, thick] (13.5,1.4) to (13.5,0.5);
        \draw[draw=base03, ->, thick] (17.25, 2.5) to (18.75, 2.5);
        \draw[draw=base03, ->, thick] (22.25, 2.5) to (23.75, 2.5);
    \end{tikzpicture}
    \caption{\label{fig:strides_subsampling} An alternative way of viewing
        strides. Instead of translating the $3 \times 3$ kernel by increments of
        $s = 2$ (left), the kernel is translated by increments of $1$ and only
        one in $s = 2$ output elements is retained (right).}
\end{figure}
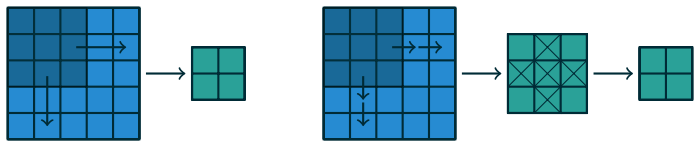

Note that strides constitute a form of \emph{subsampling}. As an alternative to
being interpreted as a measure of how much the kernel is translated, strides
can also be viewed as how much of the output is retained. For instance, moving
the kernel by hops of two is equivalent to moving the kernel by hops of one but
retaining only odd output elements (\autoref{fig:strides_subsampling}).

\begin{figure}[p]
    \centering
    \begin{tikzpicture}[scale=.35,every node/.style={minimum size=1cm}, on grid]
        \begin{scope}[xshift=0cm,yshift=0cm]
            \begin{scope}[xshift=0cm,yshift=0cm]
                \draw[draw=base03,fill=violet,thick]
                    (0,0) grid (5,5) rectangle (0,0);
            \end{scope}
            \begin{scope}[xshift=0.5cm,yshift=0.5cm]
                \draw[draw=base03,fill=blue,thick]
                    (0,0) grid (5,5) rectangle (0,0);
            \end{scope}
        \end{scope}
        \foreach \x in {-10,1,11} {%
            \begin{scope}[xshift=\x cm,yshift=10cm]
                \begin{scope}[xshift=0cm,yshift=0cm]
                    \draw[draw=base03,fill=violet,thick]
                        (0,0) grid (3,3) rectangle (0,0);
                \end{scope}
                \begin{scope}[xshift=0.5cm,yshift=0.5cm]
                    \draw[draw=base03,fill=blue,thick]
                        (0,0) grid (3,3) rectangle (0,0);
                \end{scope}
            \end{scope}
            \begin{scope}[xshift=\x cm,yshift=20cm]\begin{scope}[xshift=0.5cm]
                \draw[draw=base03,fill=cyan,thick]
                    (0,0) grid (3,3) rectangle (0,0);
            \end{scope}\end{scope}
        }
        \begin{scope}[xshift=1cm,yshift=30cm]
            \foreach \s in {0.0,0.5,1.0} {%
                \begin{scope}[xshift=\s cm,yshift=\s cm]
                    \draw[draw=base03,fill=cyan,thick]
                        (0,0) grid (3,3) rectangle (0,0);
                \end{scope}
            }
        \end{scope}
        \draw[->, thick] (-0.5,2.5) to (-8.5,9.5);
        \draw[->, thick] (3,6) to (3,9.5);
        \draw[->, thick] (6,3.5) to (12.5,9.5);
        \draw[thick]  (-8,14.5) to (-8,16);
        \draw[->, thick]  (-8,18) to (-8,19.5);
        \node[thick] (p1) at (-8,17) {$+$};
        \draw[thick]  (3,14.5) to (3,16);
        \draw[->, thick]  (3,18) to (3,19.5);
        \node[thick] (p2) at (3,17) {$+$};
        \draw[thick]  (13,14.5) to (13,16);
        \draw[->, thick]  (13,18) to (13,19.5);
        \node[thick] (p3) at (13,17) {$+$};
        \draw[->, thick]  (-8,23.5) to (2,29.5);
        \draw[->, thick]  (3,23.5) to (2.5,29.5);
        \draw[->, thick]  (13,23.5) to (3,29.5);
    \end{tikzpicture}
    \caption{\label{fig:full_picture} A convolution mapping from two input
        feature maps to three output feature maps using a $3 \times 2 \times 3
        \times 3$ collection of kernels $\mathbf{w}$. In the left pathway, input
        feature map 1 is convolved with kernel $\mathbf{w}_{1,1}$ and input
        feature map 2 is convolved with kernel $\mathbf{w}_{1,2}$, and the
        results are summed together elementwise to form the first output feature
        map. The same is repeated for the middle and right pathways to form the
        second and third feature maps, and all three output feature maps are
        grouped together to form the output.}
\end{figure}
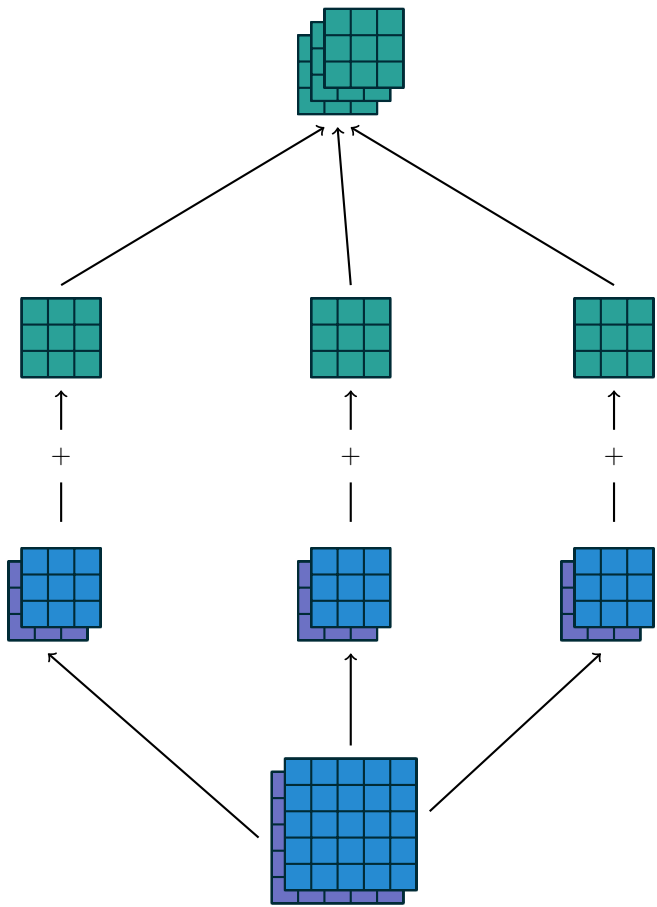

\section{Pooling}

In addition to discrete convolutions themselves, {\em pooling\/} operations
make up another important building block in CNNs. Pooling operations reduce
the size of feature maps by using some function to summarize subregions, such
as taking the average or the maximum value.

\begin{figure}[p]
    \centering
    \includegraphics[width=0.32\textwidth]{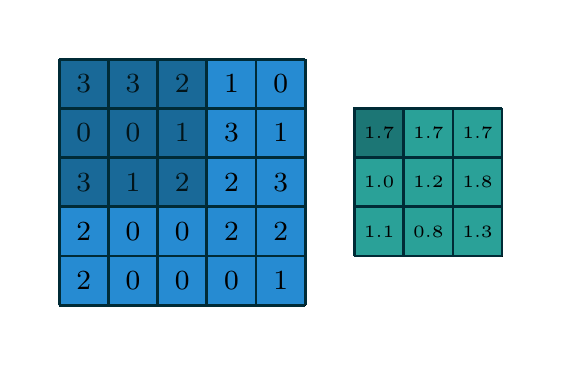}
    \includegraphics[width=0.32\textwidth]{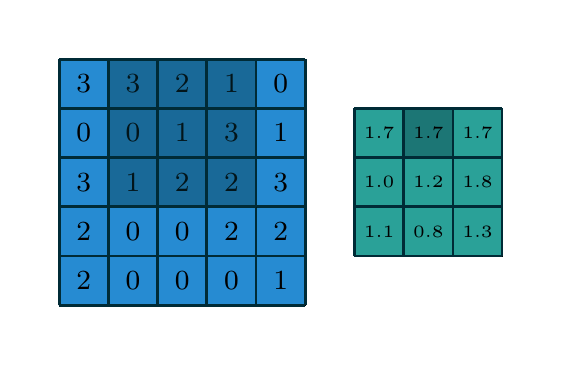}
    \includegraphics[width=0.32\textwidth]{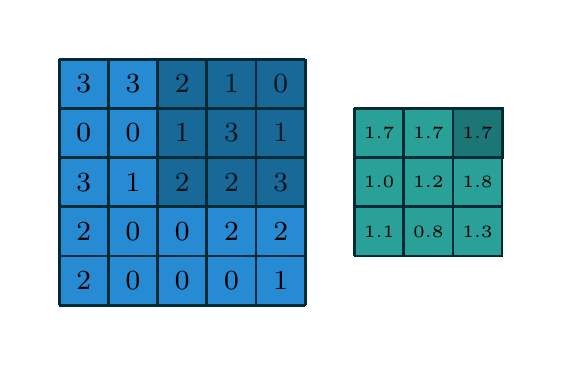}
    \includegraphics[width=0.32\textwidth]{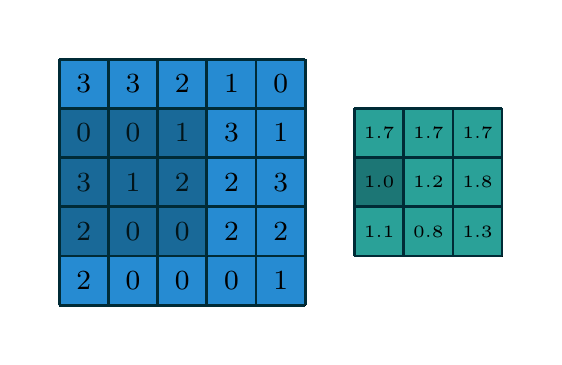}
    \includegraphics[width=0.32\textwidth]{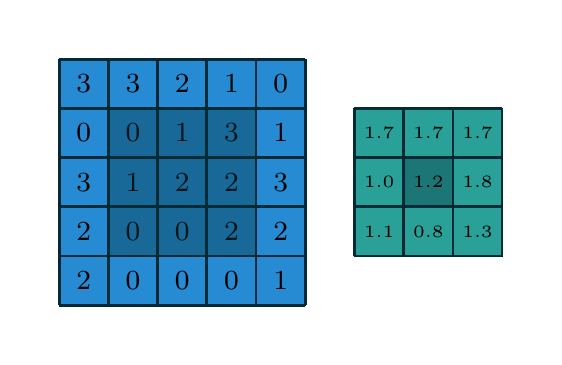}
    \includegraphics[width=0.32\textwidth]{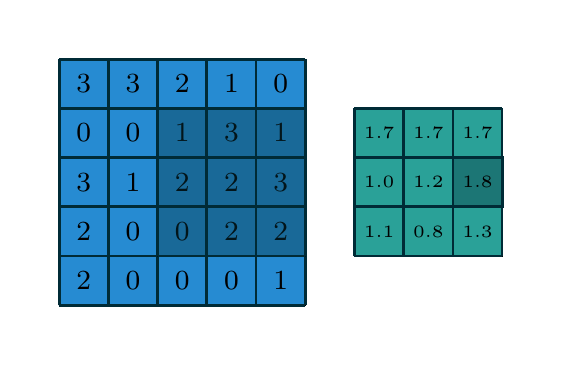}
    \includegraphics[width=0.32\textwidth]{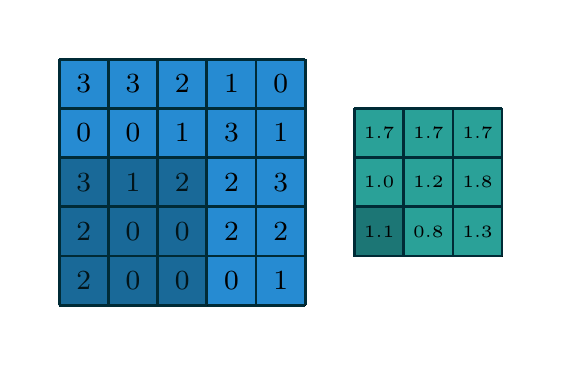}
    \includegraphics[width=0.32\textwidth]{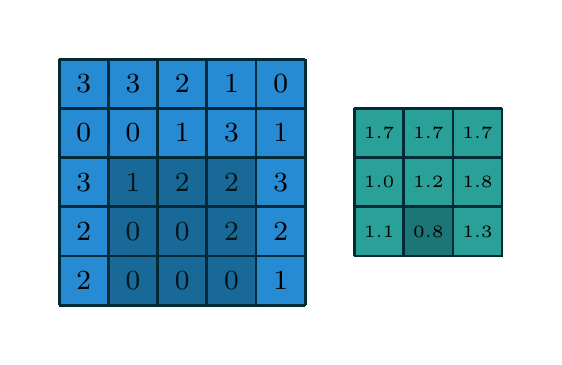}
    \includegraphics[width=0.32\textwidth]{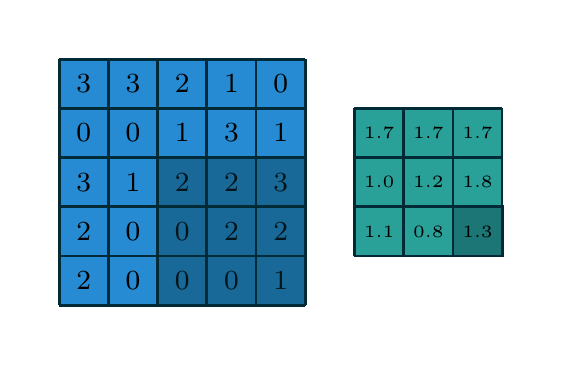}
    \caption{\label{fig:numerical_average_pooling} Computing the output values
        of a $3 \times 3$ average pooling operation on a $5 \times 5$ input
        using $1 \times 1$ strides.}
\end{figure}

\begin{figure}[p]
    \centering
    \includegraphics[width=0.32\textwidth]{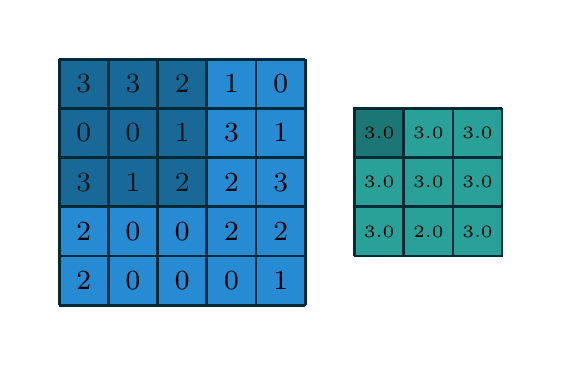}
    \includegraphics[width=0.32\textwidth]{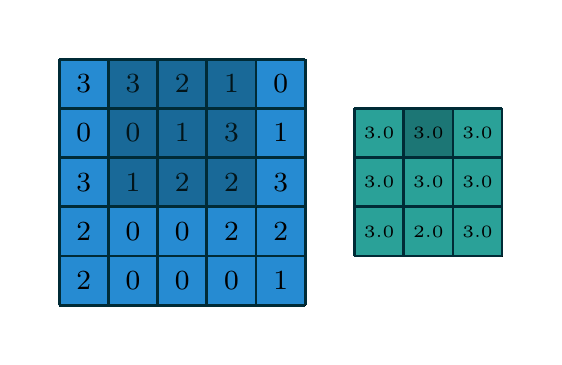}
    \includegraphics[width=0.32\textwidth]{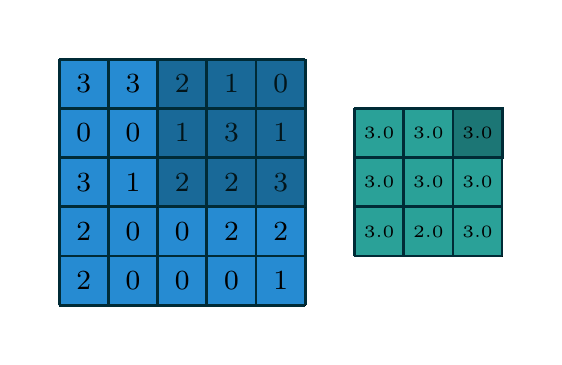}
    \includegraphics[width=0.32\textwidth]{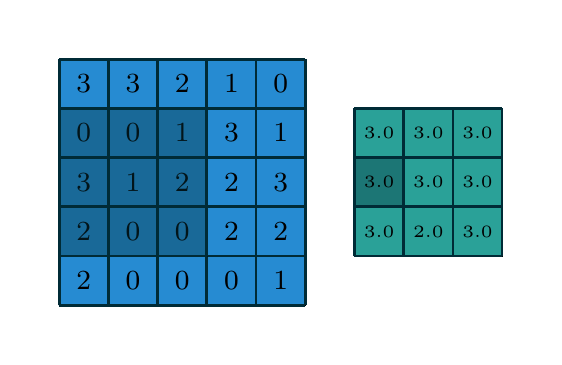}
    \includegraphics[width=0.32\textwidth]{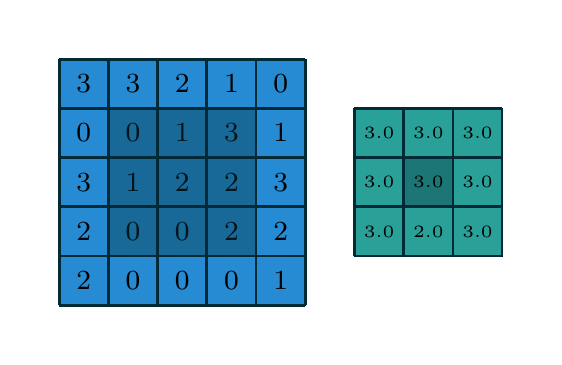}
    \includegraphics[width=0.32\textwidth]{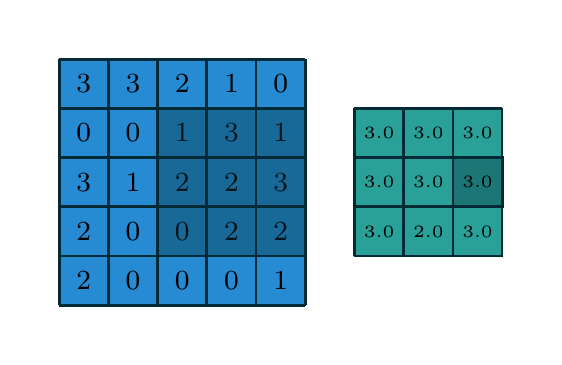}
    \includegraphics[width=0.32\textwidth]{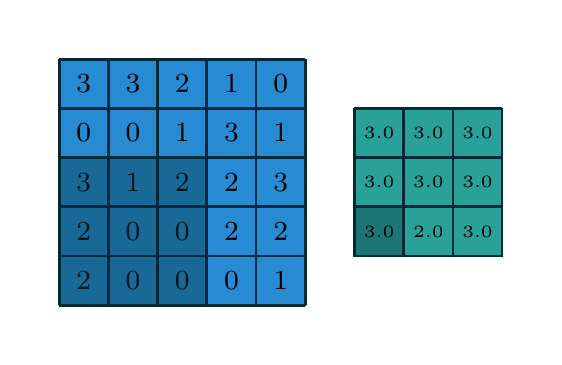}
    \includegraphics[width=0.32\textwidth]{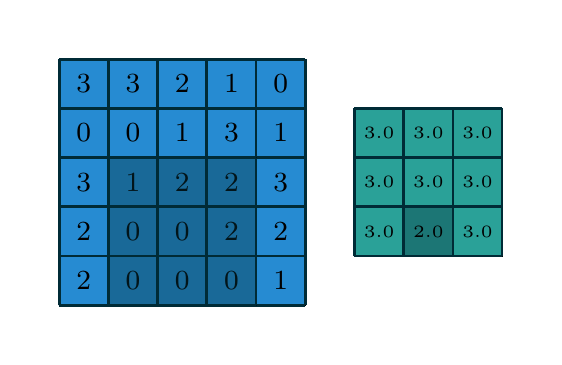}
    \includegraphics[width=0.32\textwidth]{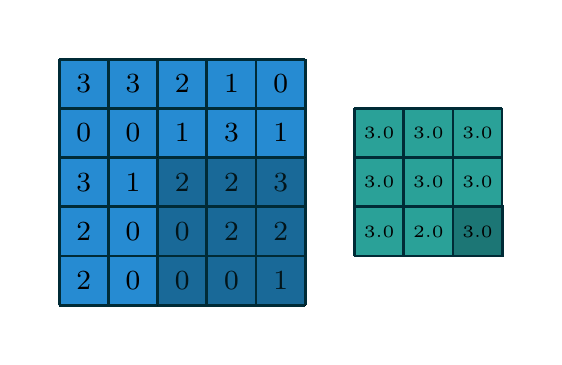}
    \caption{\label{fig:numerical_max_pooling} Computing the output values of a
        $3 \times 3$ max pooling operation on a $5 \times 5$ input using $1
        \times 1$ strides.}
\end{figure}

Pooling works by sliding a window across the input and feeding the content of
the window to a {\em pooling function}. In some sense, pooling works very much
like a discrete convolution, but replaces the linear combination described by
the kernel with some other function. \autoref{fig:numerical_average_pooling}
provides an example for average pooling, and \autoref{fig:numerical_max_pooling}
does the same for max pooling.

The following properties affect the output size $o_j$ of a pooling layer
along axis $j$:

\begin{itemize}
    \item $i_j$: input size along axis $j$,
    \item $k_j$: pooling window size along axis $j$,
    \item $s_j$: stride (distance between two consecutive positions of the
        pooling window) along axis $j$.
\end{itemize}

\chapter{Convolution arithmetic}

The analysis of the relationship between convolutional layer properties is eased
by the fact that they don't interact across axes, i.e., the choice of kernel
size, stride and zero padding along axis $j$ only affects the output size of
axis $j$. Because of that, this chapter will focus on the following simplified
setting:

\begin{itemize}
    \item 2-D discrete convolutions ($N = 2$),
    \item square inputs ($i_1 = i_2 = i$),
    \item square kernel size ($k_1 = k_2 = k$),
    \item same strides along both axes ($s_1 = s_2 = s$),
    \item same zero padding along both axes ($p_1 = p_2 = p$).
\end{itemize}

This facilitates the analysis and the visualization, but keep in mind that the
results outlined here also generalize to the N-D and non-square cases.

\section{No zero padding, unit strides}

\begin{figure}[p]
    \centering
    \includegraphics[width=0.24\textwidth]{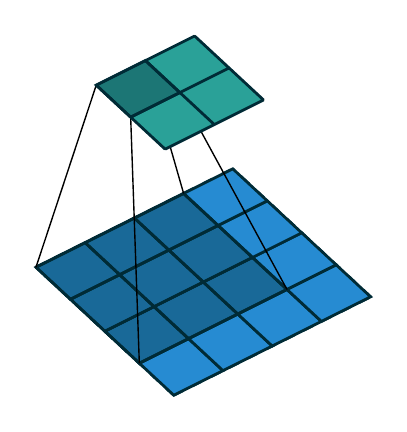}
    \includegraphics[width=0.24\textwidth]{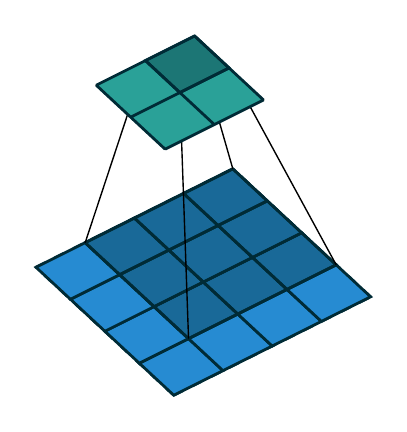}
    \includegraphics[width=0.24\textwidth]{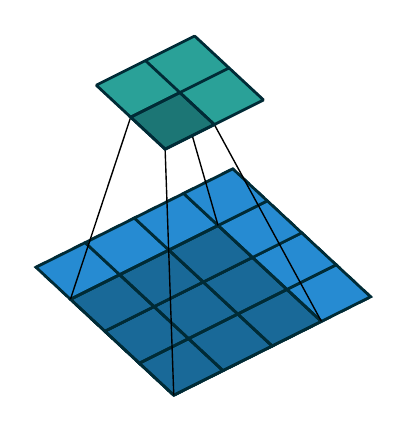}
    \includegraphics[width=0.24\textwidth]{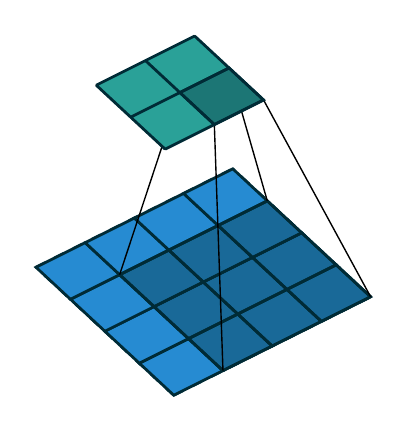}
    \caption{\label{fig:no_padding_no_strides} (No padding, unit strides)
        Convolving a $3 \times 3$ kernel over a $4 \times 4$ input using unit
        strides (i.e., $i = 4$, $k = 3$, $s = 1$ and $p = 0$).}
\end{figure}

The simplest case to analyze is when the kernel just slides across every
position of the input (i.e., $s = 1$ and $p = 0$).
\autoref{fig:no_padding_no_strides} provides an example for $i = 4$ and $k =
3$.

One way of defining the output size in this case is by the number of possible
placements of the kernel on the input. Let's consider the width axis: the kernel
starts on the leftmost part of the input feature map and slides by steps of one
until it touches the right side of the input. The size of the output will be
equal to the number of steps made, plus one, accounting for the initial position
of the kernel (\autoref{fig:no_padding_no_strides_explained}). The same logic
applies for the height axis.

More formally, the following relationship can be inferred:

\begin{relationship}\label{rel:no_padding_no_strides}
For any $i$ and $k$, and for $s = 1$ and $p = 0$,
\begin{equation*}
    o = (i - k) + 1.
\end{equation*}
\end{relationship}

\begin{figure}[p]
    \centering
    \includegraphics[width=0.24\textwidth]{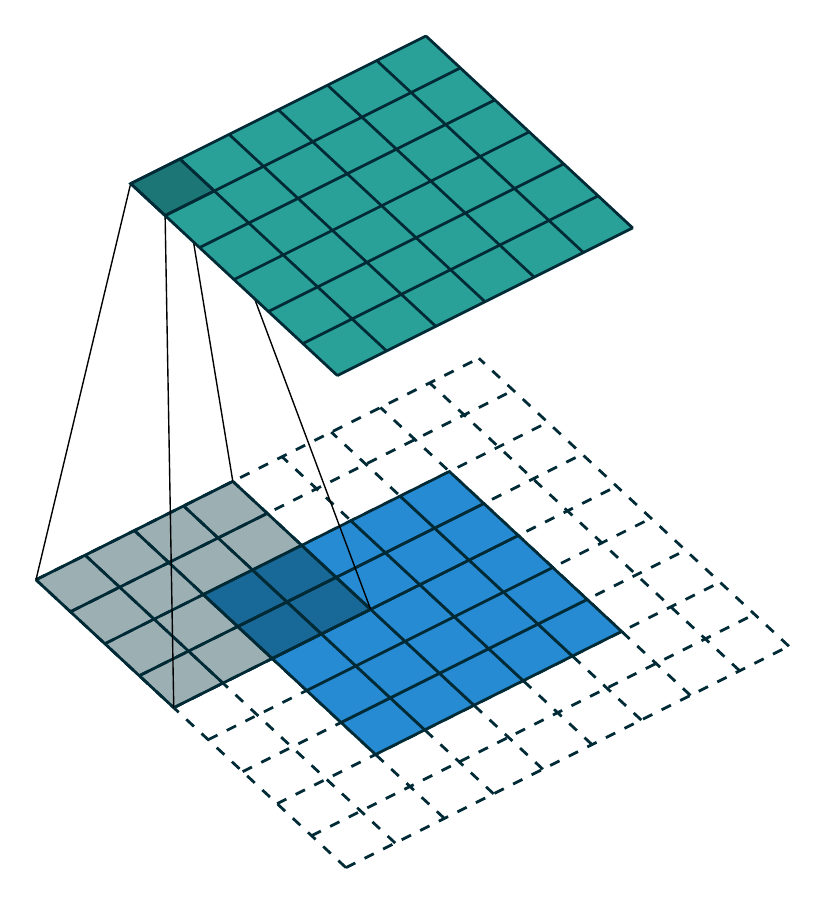}
    \includegraphics[width=0.24\textwidth]{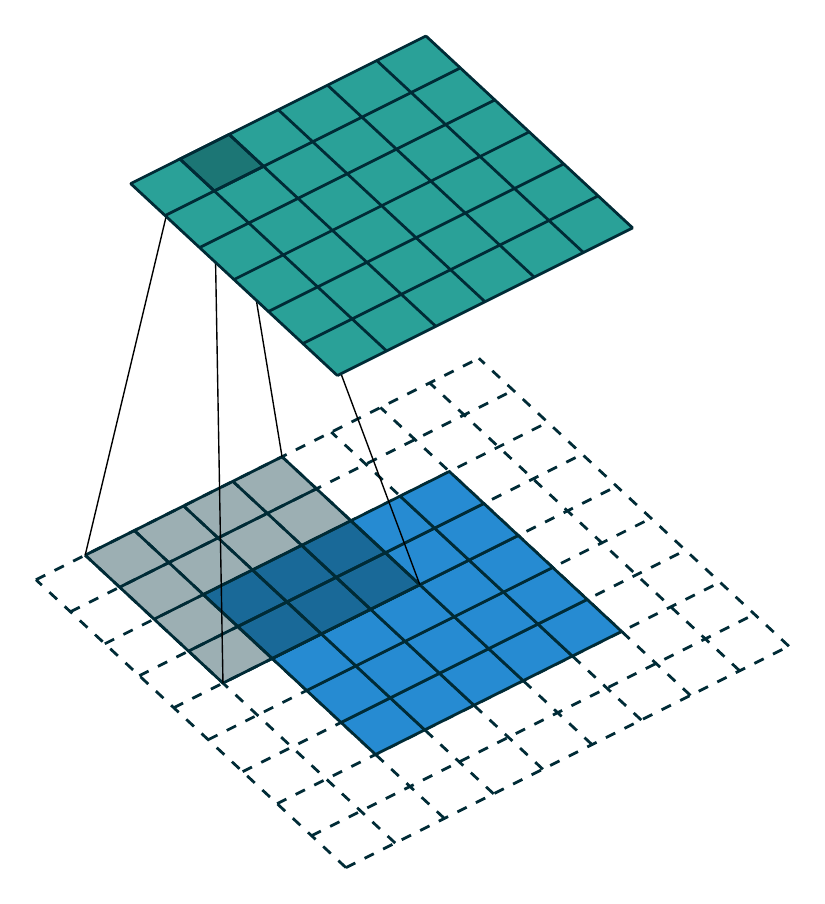}
    \includegraphics[width=0.24\textwidth]{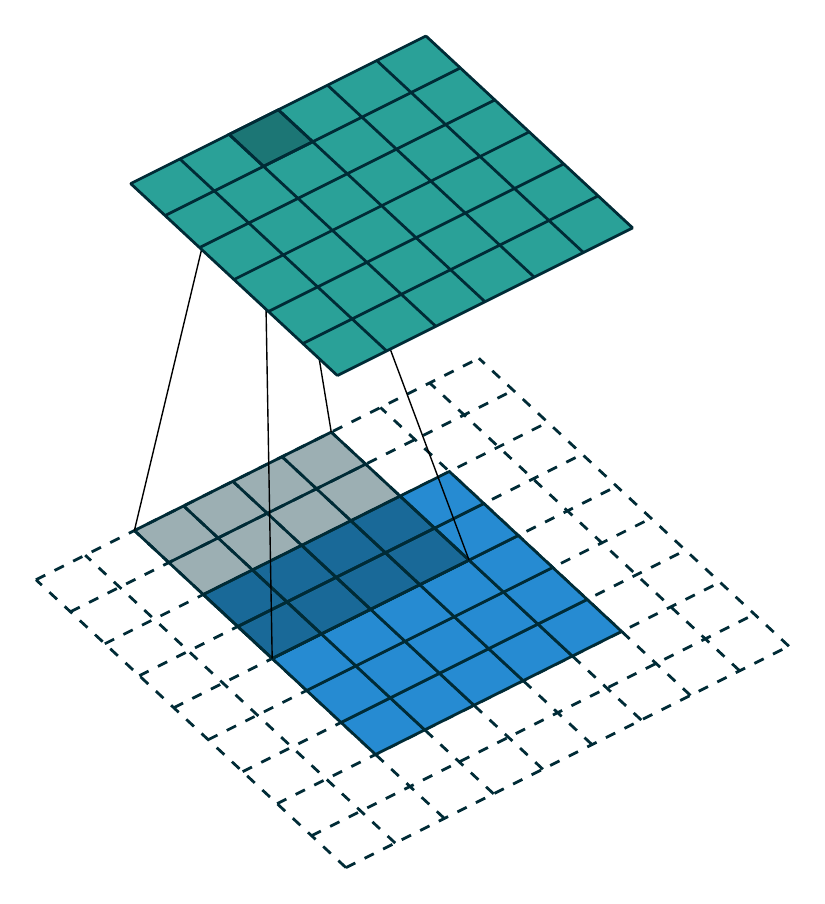}
    \includegraphics[width=0.24\textwidth]{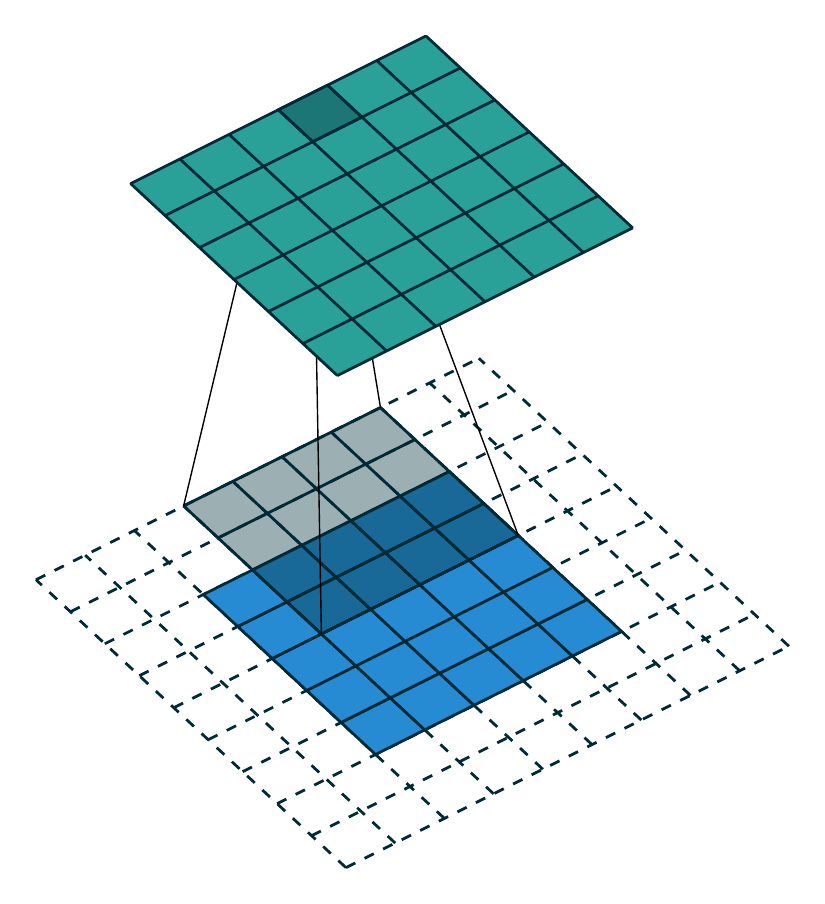}
    \caption{\label{fig:arbitrary_padding_no_strides} (Arbitrary padding, unit
        strides) Convolving a $4 \times 4$ kernel over a $5 \times 5$ input
        padded with a $2 \times 2$ border of zeros using unit strides (i.e.,
        $i = 5$, $k = 4$, $s = 1$ and $p = 2$).}
\end{figure}

\begin{figure}[p]
    \centering
    \includegraphics[width=0.24\textwidth]{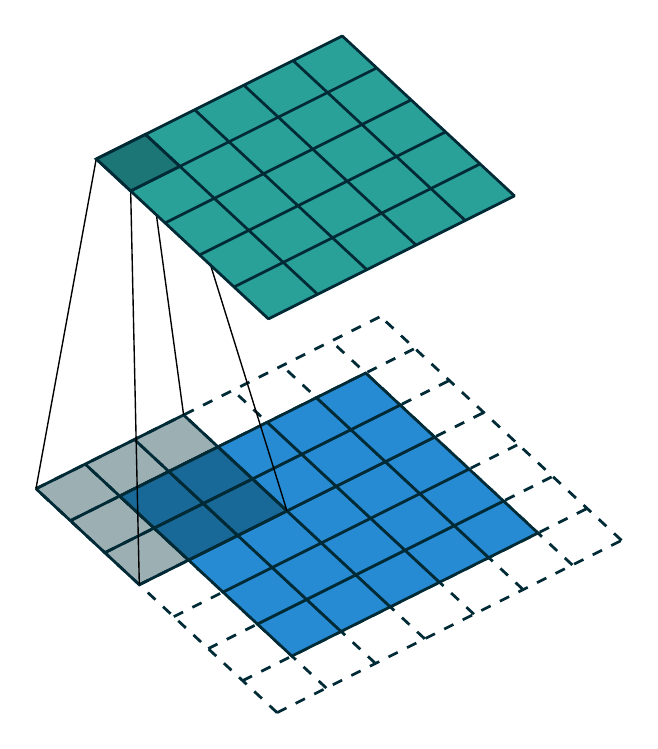}
    \includegraphics[width=0.24\textwidth]{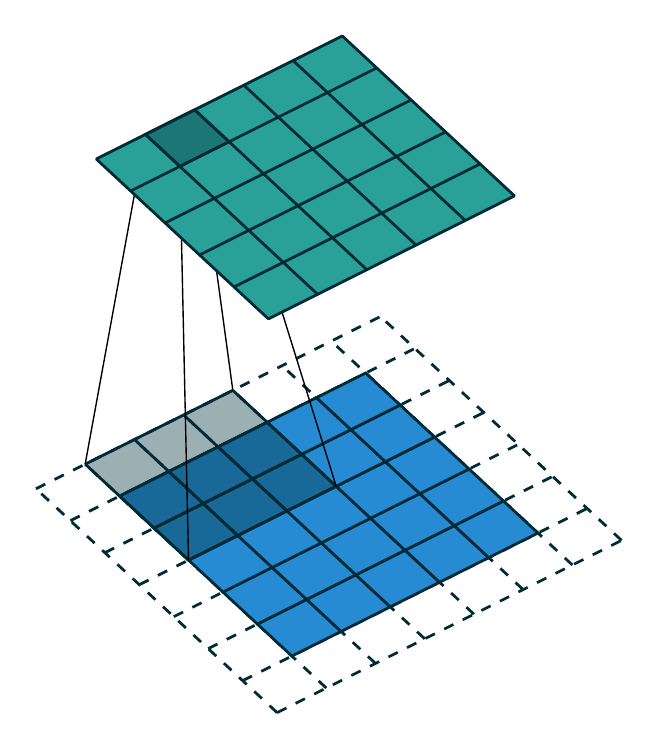}
    \includegraphics[width=0.24\textwidth]{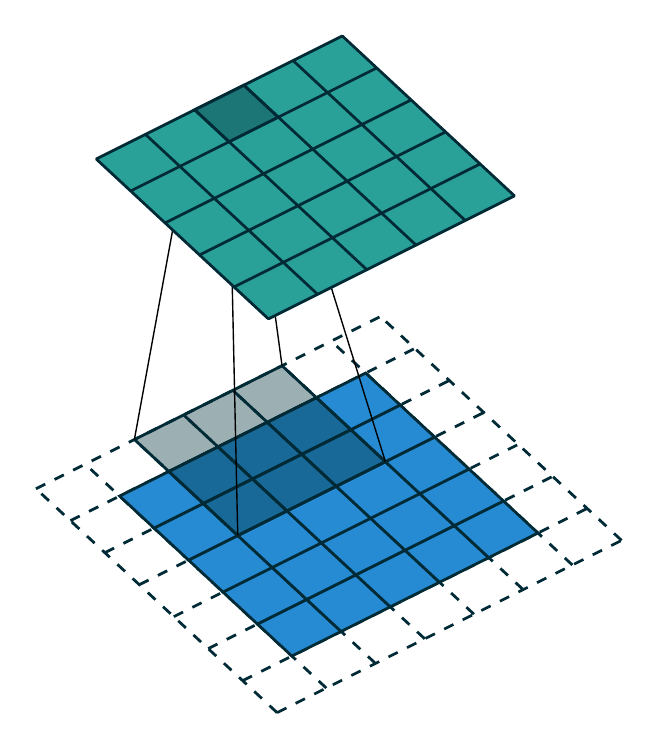}
    \includegraphics[width=0.24\textwidth]{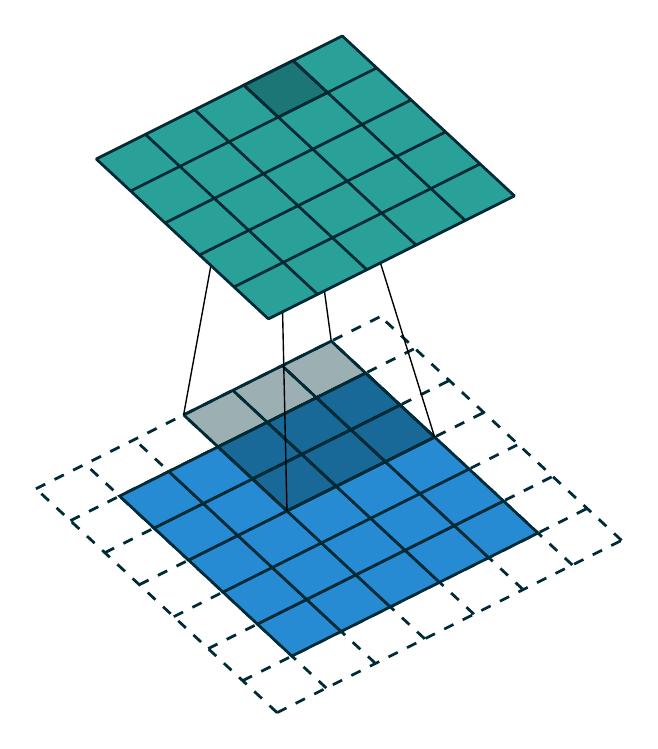}
    \caption{\label{fig:same_padding_no_strides} (Half padding, unit strides)
        Convolving a $3 \times 3$ kernel over a $5 \times 5$ input using half
        padding and unit strides (i.e., $i = 5$, $k = 3$, $s = 1$ and $p = 1$).}
\end{figure}

\begin{figure}[p]
    \centering
    \includegraphics[width=0.24\textwidth]{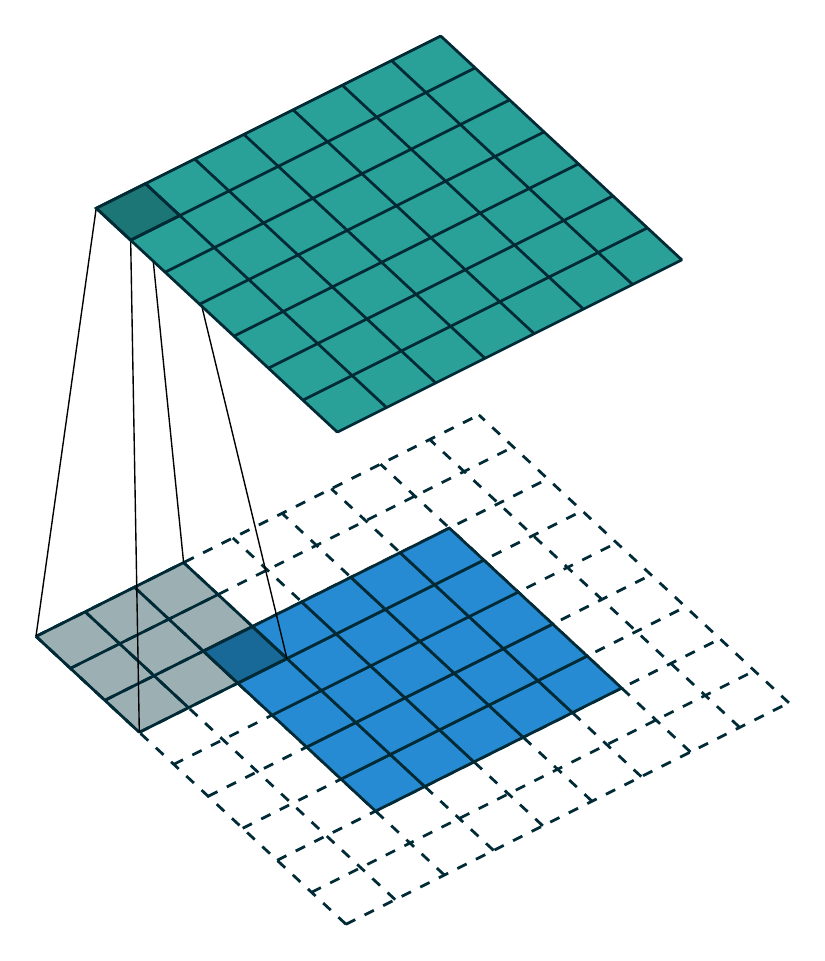}
    \includegraphics[width=0.24\textwidth]{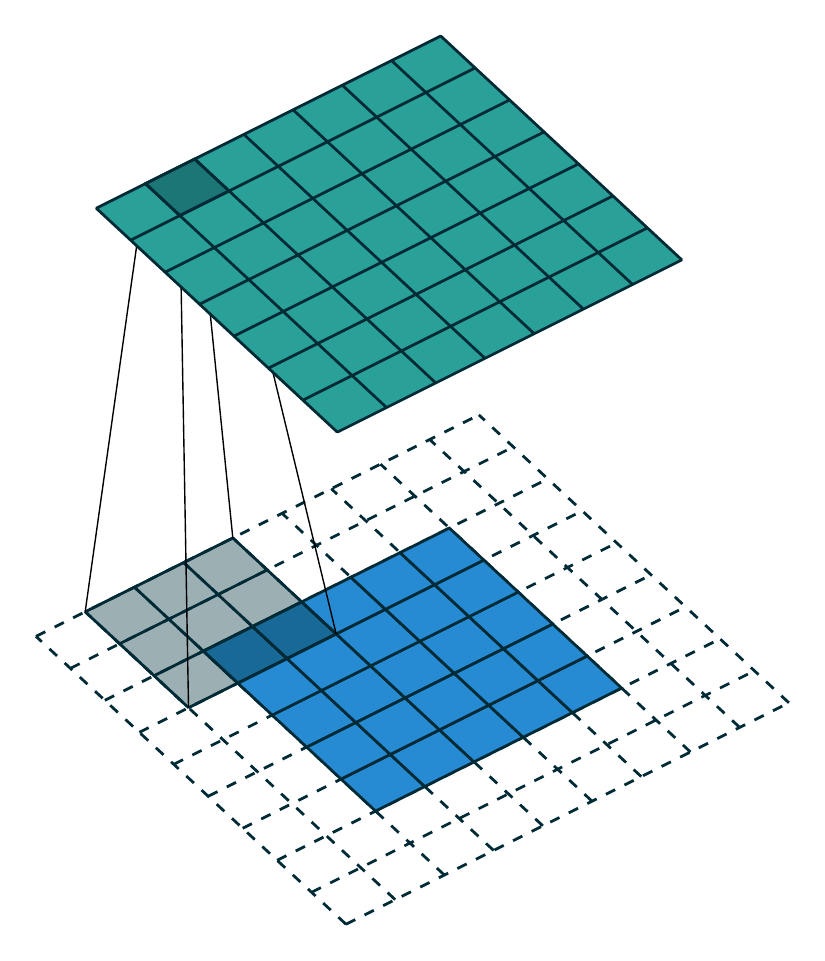}
    \includegraphics[width=0.24\textwidth]{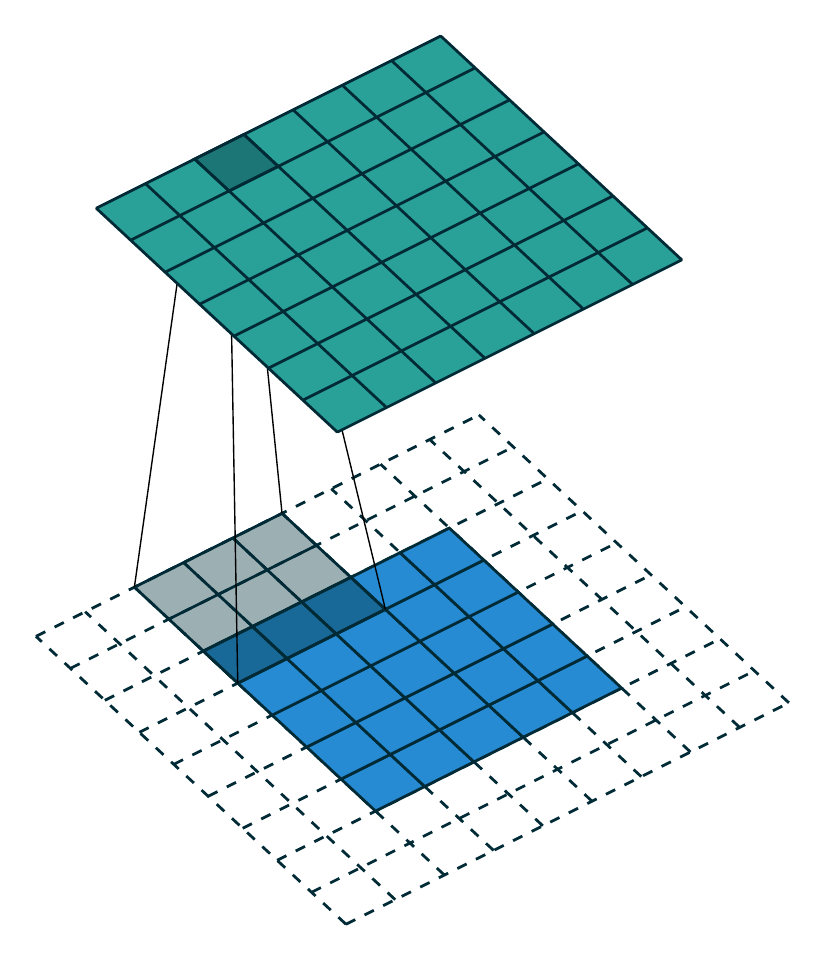}
    \includegraphics[width=0.24\textwidth]{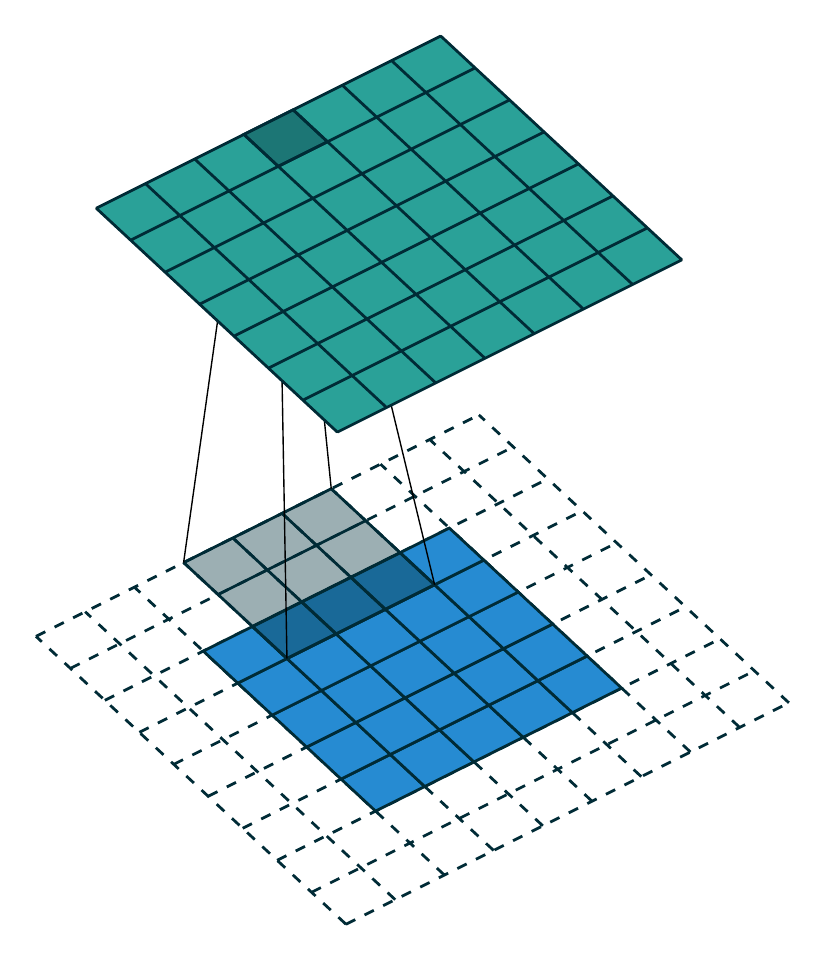}
    \caption{\label{fig:full_padding_no_strides} (Full padding, unit strides)
        Convolving a $3 \times 3$ kernel over a $5 \times 5$ input using full
        padding and unit strides (i.e., $i = 5$, $k = 3$, $s = 1$ and $p = 2$).}
\end{figure}

\section{Zero padding, unit strides}

To factor in zero padding (i.e., only restricting to $s = 1$), let's consider
its effect on the effective input size: padding with $p$ zeros changes the
effective input size from $i$ to $i + 2p$. In the general case,
\autoref{rel:no_padding_no_strides} can then be used to infer the following
relationship:

\begin{relationship}\label{rel:arbitrary_padding_no_strides}
For any $i$, $k$ and $p$, and for $s = 1$,
\begin{equation*}
    o = (i - k) + 2p + 1.
\end{equation*}
\end{relationship}

\noindent \autoref{fig:arbitrary_padding_no_strides} provides an example for $i
= 5$, $k = 4$ and $p = 2$.

In practice, two specific instances of zero padding are used quite extensively
because of their respective properties. Let's discuss them in more detail.

\subsection{Half (same) padding}

Having the output size be the same as the input size (i.e., $o = i$) can be a
desirable property:

\begin{relationship}\label{rel:same_padding_no_strides}
For any $i$ and for $k$ odd ($k = 2n + 1, \quad n \in \mathbb{N}$), $s = 1$ and
$p = \lfloor k / 2 \rfloor = n$,
\begin{equation*}
\begin{split}
    o &= i + 2 \lfloor k / 2 \rfloor - (k - 1) \\
      &= i + 2n - 2n \\
      &= i.
\end{split}
\end{equation*}
\end{relationship}

\noindent This is sometimes referred to as {\em half\/} (or {\em same\/})
padding. \autoref{fig:same_padding_no_strides} provides an example for
$i = 5$, $k = 3$ and (therefore) $p = 1$.

\subsection{Full padding}

While convolving a kernel generally {\em decreases\/} the output size with
respect to the input size, sometimes the opposite is required. This can be
achieved with proper zero padding:

\begin{relationship}\label{rel:full_padding_no_strides}
For any $i$ and $k$, and for $p = k - 1$ and $s = 1$,
\begin{equation*}
\begin{split}
    o &= i + 2(k - 1) - (k - 1) \\
      &= i + (k - 1).
\end{split}
\end{equation*}
\end{relationship}

\noindent This is sometimes referred to as {\em full\/} padding, because in this
setting every possible partial or complete superimposition of the kernel on the
input feature map is taken into account. \autoref{fig:full_padding_no_strides}
provides an example for $i = 5$, $k = 3$ and (therefore) $p = 2$.

\section{No zero padding, non-unit strides}

\begin{figure}[p]
    \centering
    \includegraphics[width=0.24\textwidth]{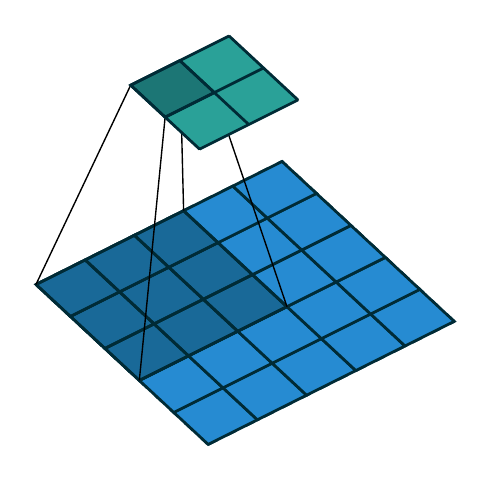}
    \includegraphics[width=0.24\textwidth]{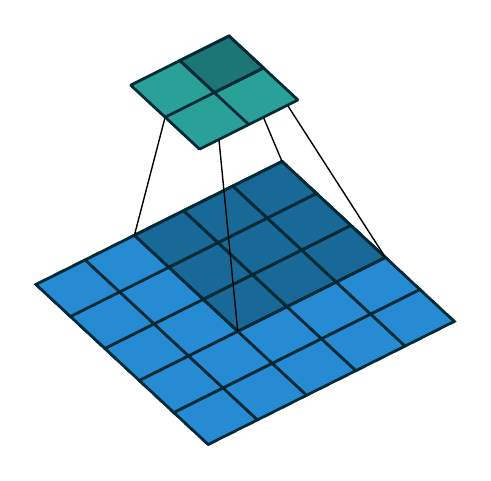}
    \includegraphics[width=0.24\textwidth]{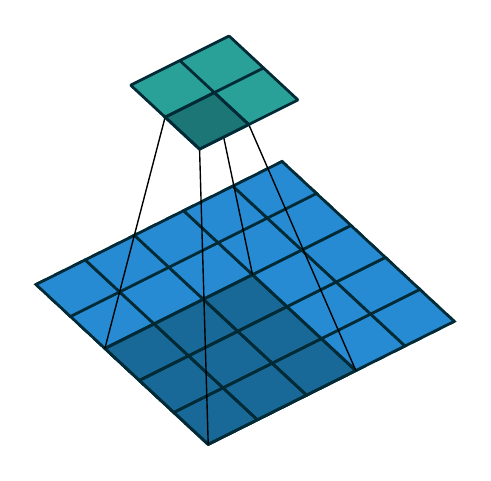}
    \includegraphics[width=0.24\textwidth]{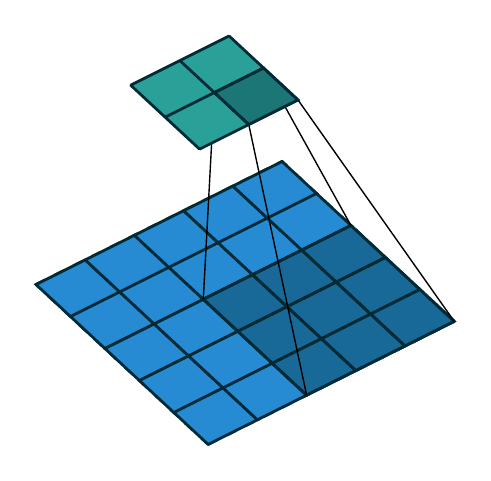}
    \caption{\label{fig:no_padding_strides} (No zero padding, arbitrary
        strides) Convolving a $3 \times 3$ kernel over a $5 \times 5$ input
        using $2 \times 2$ strides (i.e., $i = 5$, $k = 3$, $s = 2$ and
        $p = 0$).}
\end{figure}

All relationships derived so far only apply for unit-strided convolutions.
Incorporating non unitary strides requires another inference leap. To
facilitate the analysis, let's momentarily ignore zero padding (i.e., $s > 1$
and $p = 0$). \autoref{fig:no_padding_strides} provides an example for $i =
5$, $k = 3$ and $s = 2$.

Once again, the output size can be defined in terms of the number of possible
placements of the kernel on the input. Let's consider the width axis: the
kernel starts as usual on the leftmost part of the input, but this time it
slides by steps of size $s$ until it touches the right side of the input. The
size of the output is again equal to the number of steps made, plus one,
accounting for the initial position of the kernel
(\autoref{fig:no_padding_strides_explained}). The same logic applies for the
height axis.

From this, the following relationship can be inferred:

\begin{relationship}\label{rel:no_padding_strides}
For any $i$, $k$ and $s$, and for $p = 0$,
\begin{equation*}
    o = \left\lfloor \frac{i - k}{s} \right\rfloor + 1.
\end{equation*}
\end{relationship}

\noindent The floor function accounts for the fact that sometimes the last
possible step does {\em not\/} coincide with the kernel reaching the end of the
input, i.e., some input units are left out (see
\autoref{fig:padding_strides_odd} for an example of such a case).

\section{Zero padding, non-unit strides}

The most general case (convolving over a zero padded input using non-unit
strides) can be derived by applying \autoref{rel:no_padding_strides} on an
effective input of size $i + 2p$, in analogy to what was done for
\autoref{rel:arbitrary_padding_no_strides}:

\begin{relationship}\label{rel:padding_strides}
For any $i$, $k$, $p$ and $s$,
\begin{equation*}
    o = \left\lfloor \frac{i + 2p - k}{s} \right\rfloor + 1.
\end{equation*}
\end{relationship}

\noindent As before, the floor function means that in some cases a convolution
will produce the same output size for multiple input sizes. More specifically,
if $i + 2p - k$ is a multiple of $s$, then any input size $j = i + a, \quad a
\in \{0,\ldots,s - 1\}$ will produce the same output size. Note that this
ambiguity applies only for $s > 1$.

\begin{figure}[p]
    \centering
    \includegraphics[width=0.24\textwidth]{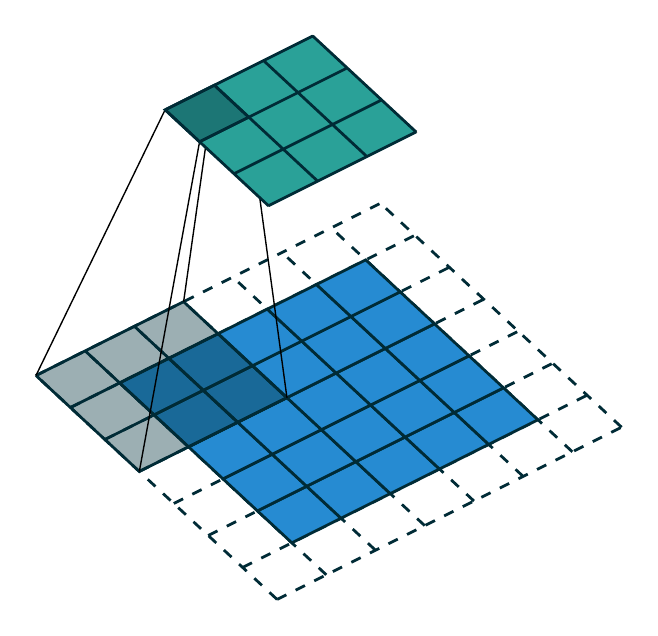}
    \includegraphics[width=0.24\textwidth]{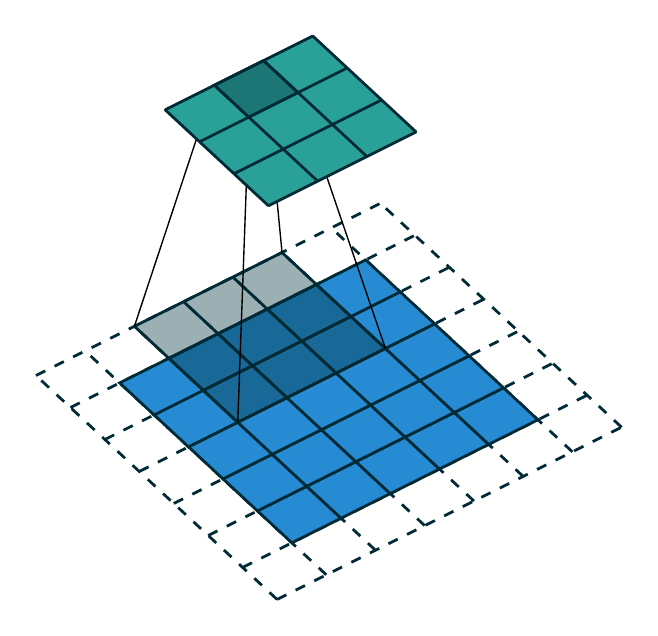}
    \includegraphics[width=0.24\textwidth]{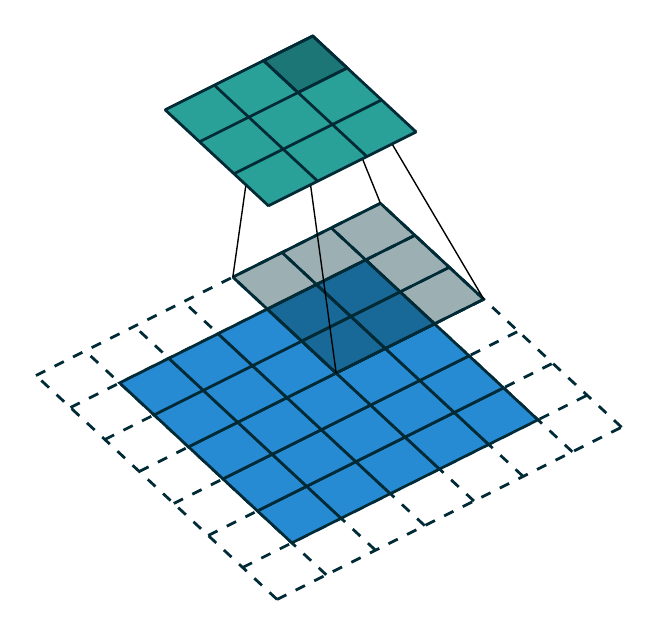}
    \includegraphics[width=0.24\textwidth]{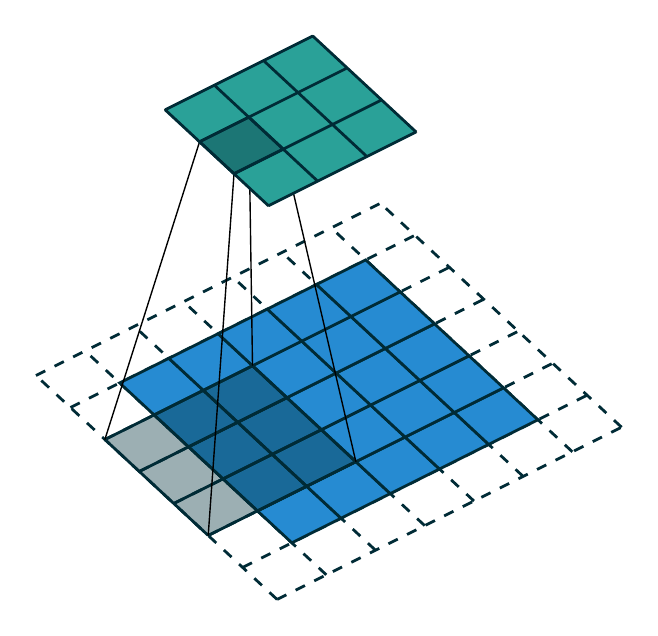}
    \caption{\label{fig:padding_strides} (Arbitrary padding and strides)
        Convolving a $3 \times 3$ kernel over a $5 \times 5$ input padded with
        a $1 \times 1$ border of zeros using $2 \times 2$ strides (i.e.,
        $i = 5$, $k = 3$, $s = 2$ and $p = 1$).}
\end{figure}

\begin{figure}[p]
    \centering
    \includegraphics[width=0.24\textwidth]{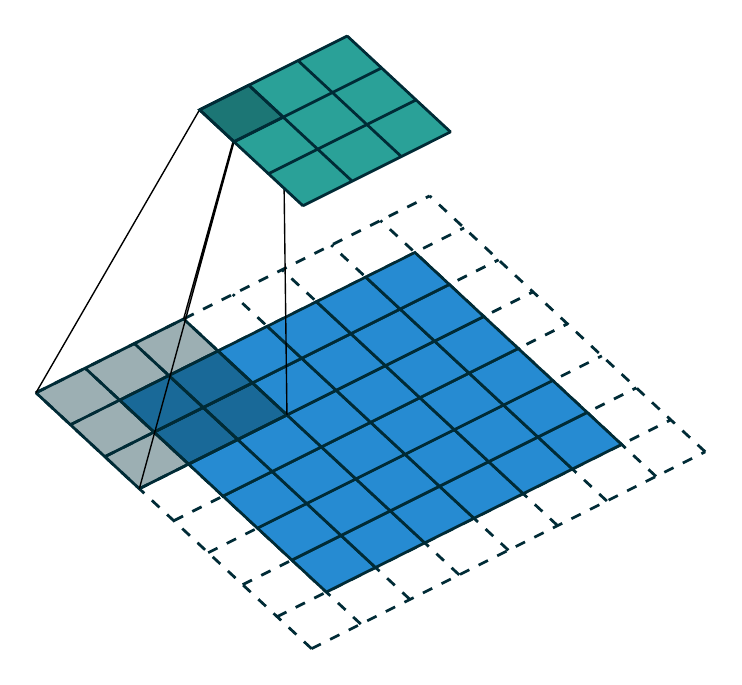}
    \includegraphics[width=0.24\textwidth]{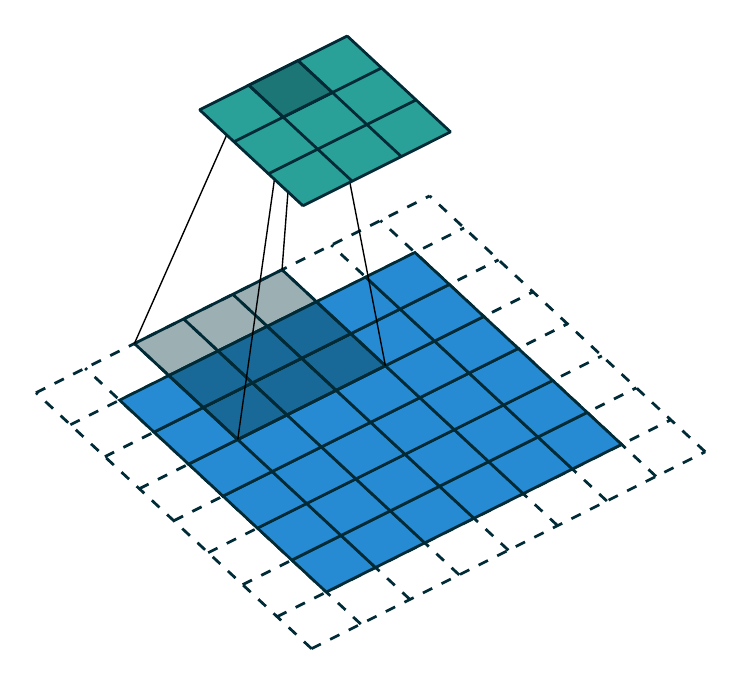}
    \includegraphics[width=0.24\textwidth]{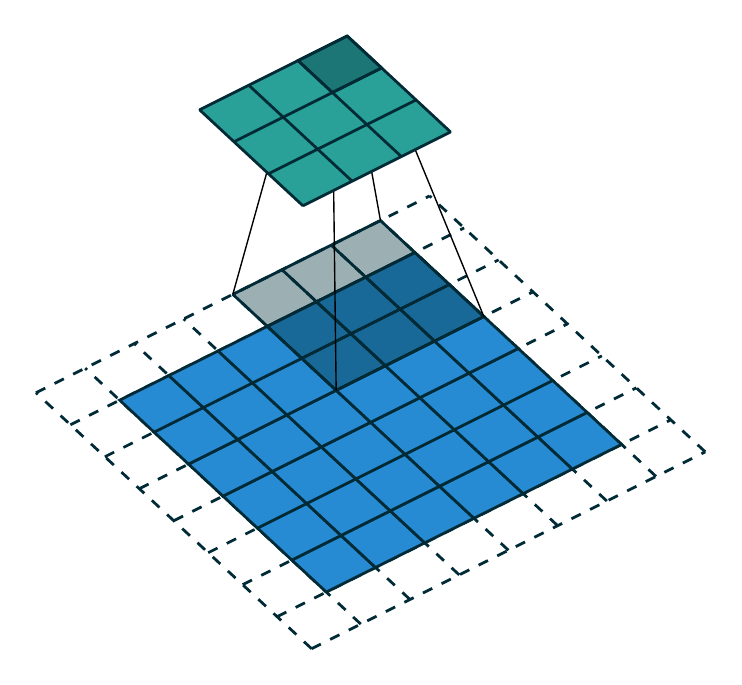}
    \includegraphics[width=0.24\textwidth]{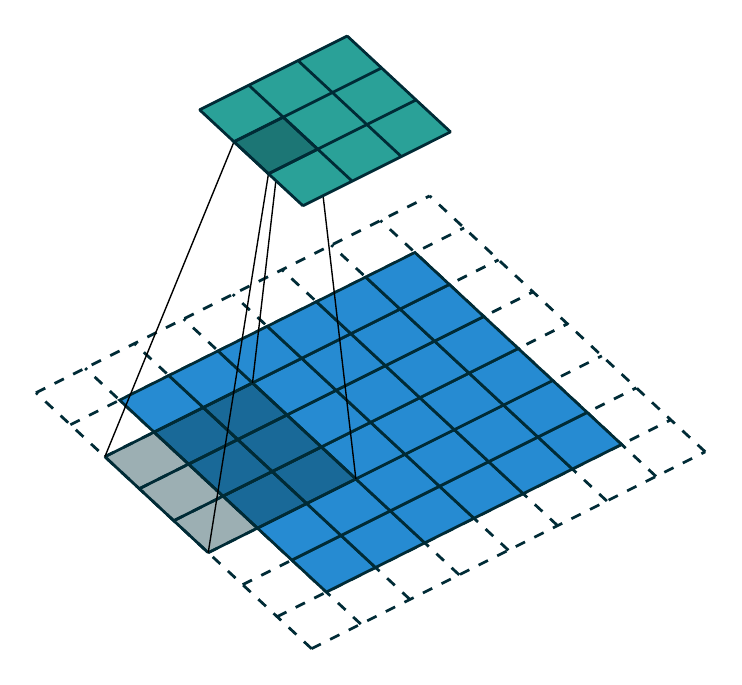}
    \caption{\label{fig:padding_strides_odd} (Arbitrary padding and strides)
        Convolving a $3 \times 3$ kernel over a $6 \times 6$ input padded with
        a $1 \times 1$ border of zeros using $2 \times 2$ strides (i.e.,
        $i = 6$, $k = 3$, $s = 2$ and $p = 1$). In this case, the bottom row
        and right column of the zero padded input are not covered by the
        kernel.}
\end{figure}

\autoref{fig:padding_strides} shows an example with $i = 5$, $k = 3$, $s = 2$
and $p = 1$, while \autoref{fig:padding_strides_odd} provides an example for
$i = 6$, $k = 3$, $s = 2$ and $p = 1$. Interestingly, despite having different
input sizes these convolutions share the same output size. While this doesn't
affect the analysis for {\em convolutions}, this will complicate the analysis
in the case of {\em transposed convolutions}.

\begin{figure}[p]
    \centering
    \begin{subfigure}[t]{0.48\textwidth}
        \centering
        \begin{tikzpicture}[scale=.35,every node/.style={minimum size=1cm},
                            on grid]
            \draw[fill=blue] (0,0) rectangle (5,5);
            \draw[draw=base03, thick] (0,0) grid (5,5);
            \draw[fill=base02, opacity=0.4] (0,2) rectangle (3,5);
            \draw[step=10mm, base03, thick] (0,2) grid (3,5);
            \draw[draw=base03, ->, thick] (2.6,3.5) to  (3.5,3.5);
            \draw[draw=base03, ->, thick] (3.6,3.5) to  (4.5,3.5);
            \draw[draw=base03, ->, thick] (1.5,2.4) to  (1.5,1.5);
            \draw[draw=base03, ->, thick] (1.5,1.4) to  (1.5,0.5);
        \end{tikzpicture}
        \caption{\label{fig:no_padding_no_strides_explained} The kernel has to
            slide two steps to the right to touch the right side of the input
            (and equivalently downwards).  Adding one to account for the
            initial kernel position, the output size is $3 \times 3$.}
    \end{subfigure}
    ~
    \begin{subfigure}[t]{0.48\textwidth}
        \centering
        \begin{tikzpicture}[scale=.35,every node/.style={minimum size=1cm},
                            on grid]
            \draw[fill=blue] (0,0) rectangle (5,5);
            \draw[draw=base03, thick] (0,0) grid (5,5);
            \draw[fill=base02, opacity=0.4] (0,2) rectangle (3,5);
            \draw[step=10mm, base03, thick] (0,2) grid (3,5);
            \draw[draw=base03, ->, thick] (2.5,3.5) to  (4.5,3.5);
            \draw[draw=base03, ->, thick] (1.5,2.5) to  (1.5,0.5);
        \end{tikzpicture}
        \caption{\label{fig:no_padding_strides_explained} The kernel has to
            slide one step of size two to the right to touch the right side of
            the input (and equivalently downwards).  Adding one to account for
            the initial kernel position, the output size is $2 \times 2$.}
    \end{subfigure}
    \caption{Counting kernel positions.}
\end{figure}

\chapter{Pooling arithmetic}

In a neural network, pooling layers provide invariance to small translations of
the input. The most common kind of pooling is \emph{max pooling}, which
consists in splitting the input in (usually non-overlapping) patches and
outputting the maximum value of each patch. Other kinds of pooling exist, e.g.,
mean or average pooling, which all share the same idea of aggregating the input
locally by applying a non-linearity to the content of some patches \citep{%
boureau-cvpr-10,boureau-icml-10,boureau-iccv-11,ICML2011Saxe_551}.

Some readers may have noticed that the treatment of convolution arithmetic only
relies on the assumption that some function is repeatedly applied onto subsets
of the input. This means that the relationships derived in the previous chapter
can be reused in the case of pooling arithmetic. Since pooling does not involve
zero padding, the relationship describing the general case is as follows:

\begin{relationship}\label{rel:pooling}
For any $i$, $k$ and $s$,
\begin{equation*}
    o = \left\lfloor \frac{i - k}{s} \right\rfloor + 1.
\end{equation*}
\end{relationship}

\noindent This relationship holds for any type of pooling.

\chapter{Transposed convolution arithmetic}

The need for transposed convolutions generally arises from the desire to use a
transformation going in the opposite direction of a normal convolution, i.e.,
from something that has the shape of the output of some convolution to
something that has the shape of its input while maintaining a connectivity
pattern that is compatible with said convolution. For instance, one might use
such a transformation as the decoding layer of a convolutional autoencoder or to
project feature maps to a higher-dimensional space.

Once again, the convolutional case is considerably more complex than the
fully-connected case, which only requires to use a weight matrix whose shape
has been transposed. However, since every convolution boils down to an
efficient implementation of a matrix operation, the insights gained from the
fully-connected case are useful in solving the convolutional case.

Like for convolution arithmetic, the dissertation about transposed convolution
arithmetic is simplified by the fact that transposed convolution properties
don't interact across axes.

The chapter will focus on the following setting:

\begin{itemize}
    \item 2-D transposed convolutions ($N = 2$),
    \item square inputs ($i_1 = i_2 = i$),
    \item square kernel size ($k_1 = k_2 = k$),
    \item same strides along both axes ($s_1 = s_2 = s$),
    \item same zero padding along both axes ($p_1 = p_2 = p$).
\end{itemize}

\noindent Once again, the results outlined generalize to the N-D and non-square
cases.

\section{Convolution as a matrix operation}

Take for example the convolution represented in
\autoref{fig:no_padding_no_strides}. If the input and output were to be unrolled
into vectors from left to right, top to bottom, the convolution could be
represented as a sparse matrix $\mathbf{C}$ where the non-zero elements are the
elements $w_{i,j}$ of the kernel (with $i$ and $j$ being the row and column of
the kernel respectively):
\begin{equation*}
\resizebox{.98\hsize}{!}{$
    \begin{pmatrix}
    w_{0,0} & w_{0,1} & w_{0,2} & 0       & w_{1,0} & w_{1,1} & w_{1,2} & 0       &
    w_{2,0} & w_{2,1} & w_{2,2} & 0       & 0       & 0       & 0       & 0       \\
    0       & w_{0,0} & w_{0,1} & w_{0,2} & 0       & w_{1,0} & w_{1,1} & w_{1,2} &
    0       & w_{2,0} & w_{2,1} & w_{2,2} & 0       & 0       & 0       & 0       \\
    0       & 0       & 0       & 0       & w_{0,0} & w_{0,1} & w_{0,2} & 0       &
    w_{1,0} & w_{1,1} & w_{1,2} & 0       & w_{2,0} & w_{2,1} & w_{2,2} & 0       \\
    0       & 0       & 0       & 0       & 0       & w_{0,0} & w_{0,1} & w_{0,2} &
    0       & w_{1,0} & w_{1,1} & w_{1,2} & 0       & w_{2,0} & w_{2,1} & w_{2,2} \\
    \end{pmatrix}$}
\end{equation*}

This linear operation takes the input matrix flattened as a 16-dimensional
vector and produces a 4-dimensional vector that is later reshaped as the $2
\times 2$ output matrix.

Using this representation, the backward pass is easily obtained by transposing
$\mathbf{C}$; in other words, the error is backpropagated by multiplying the
loss with $\mathbf{C}^T$. This operation takes a 4-dimensional vector as input
and produces a 16-dimensional vector as output, and its connectivity pattern is
compatible with $\mathbf{C}$ by construction.

Notably, the kernel $\mathbf{w}$ defines both the matrices $\mathbf{C}$ and
$\mathbf{C}^T$ used for the forward and backward passes.

\section{Transposed convolution}

Let's now consider what would be required to go the other way around, i.e., map
from a 4-dimensional space to a 16-dimensional space, while keeping the
connectivity pattern of the convolution depicted in
\autoref{fig:no_padding_no_strides}. This operation is known as a {\em
transposed convolution}.

Transposed convolutions -- also called {\em fractionally strided convolutions\/}
or {\em deconvolutions\/}\footnote{The term ``deconvolution'' is sometimes used
in the literature, but we advocate against it on the grounds that a
deconvolution is mathematically defined as the inverse of a convolution, which
is different from a transposed convolution.} -- work by swapping the forward and
backward passes of a convolution. One way to put it is to note that the kernel
defines a convolution, but whether it's a direct convolution or a transposed
convolution is determined by how the forward and backward passes are computed.

For instance, although the kernel $\mathbf{w}$ defines a convolution whose
forward and backward passes are computed by multiplying with $\mathbf{C}$ and
$\mathbf{C}^T$ respectively, it {\em also\/} defines a transposed convolution
whose forward and backward passes are computed by multiplying with
$\mathbf{C}^T$ and $(\mathbf{C}^T)^T = \mathbf{C}$ respectively.\footnote{The
    transposed convolution operation can be thought of as the gradient of {\em
    some\/} convolution with respect to its input, which is usually how
    transposed convolutions are implemented in practice.}

Finally note that it is always possible to emulate a transposed convolution with
a direct convolution. The disadvantage is that it usually involves adding many
columns and rows of zeros to the input, resulting in a much less efficient
implementation.

Building on what has been introduced so far, this chapter will proceed somewhat
backwards with respect to the convolution arithmetic chapter, deriving the
properties of each transposed convolution by referring to the direct
convolution with which it shares the kernel, and defining the equivalent direct
convolution.

\section{No zero padding, unit strides, transposed}

The simplest way to think about a transposed convolution on a given input is to
imagine such an input as being the result of a direct convolution applied on
some initial feature map. The trasposed convolution can be then considered as
the operation that allows to recover the \emph{shape}~\footnote{Note that the
  transposed convolution does not guarantee to recover the input itself, as it
  is not defined as the inverse of the convolution, but rather just returns a
  feature map that has the same width and height.} of this initial feature map.

Let's consider the convolution of a $3 \times 3$ kernel on a $4 \times 4$
input with unitary stride and no padding (i.e., $i = 4$, $k = 3$, $s = 1$ and
$p = 0$). As depicted in \autoref{fig:no_padding_no_strides}, this produces a
$2 \times 2$ output. The transpose of this convolution will then have an output
of shape $4 \times 4$ when applied on a $2 \times 2$ input.

\begin{figure}[p]
    \centering
    \includegraphics[width=0.24\textwidth]{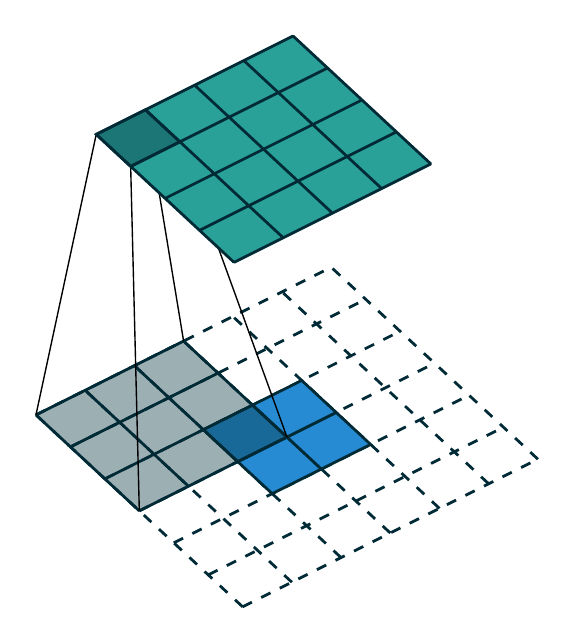}
    \includegraphics[width=0.24\textwidth]{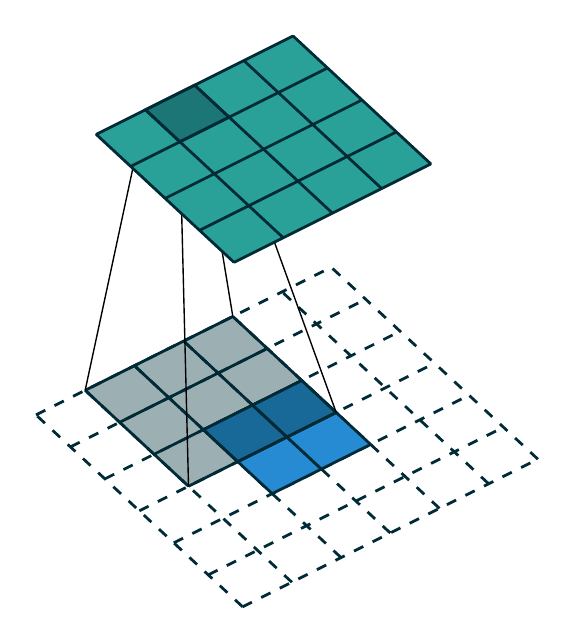}
    \includegraphics[width=0.24\textwidth]{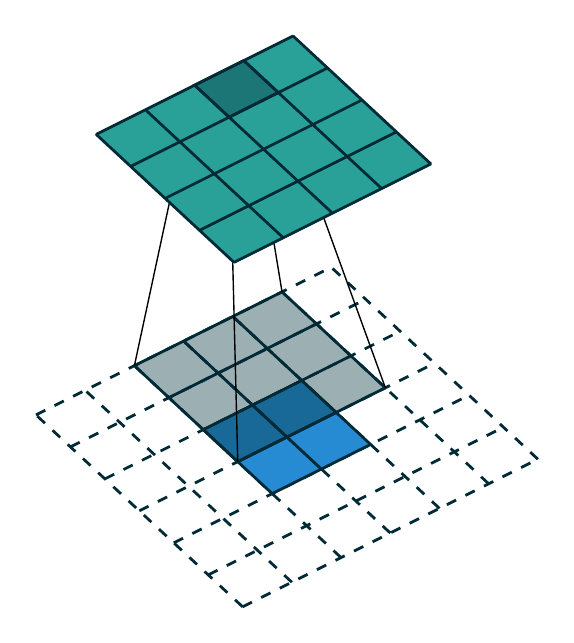}
    \includegraphics[width=0.24\textwidth]{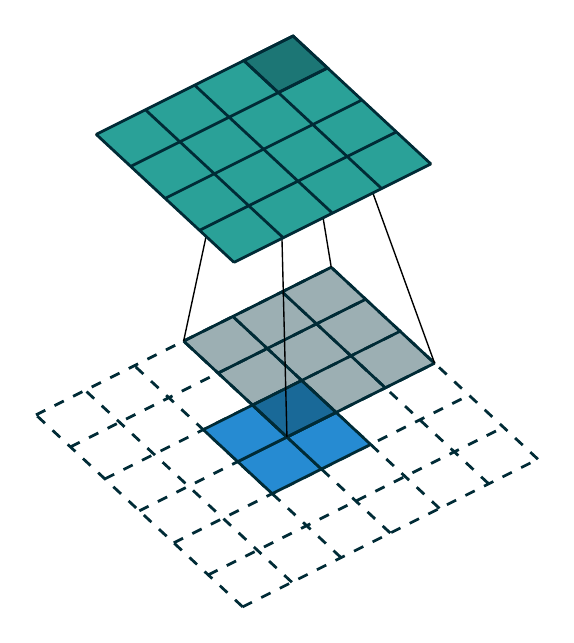}
    \caption{\label{fig:no_padding_no_strides_transposed} The transpose of
        convolving a $3 \times 3$ kernel over a $4 \times 4$ input using unit
        strides (i.e., $i = 4$, $k = 3$, $s = 1$ and $p = 0$). It is equivalent
        to convolving a $3 \times 3$ kernel over a $2 \times 2$ input padded
        with a $2 \times 2$ border of zeros using unit strides (i.e., $i' = 2$,
        $k' = k$, $s' = 1$ and $p' = 2$).}
\end{figure}

Another way to obtain the result of a transposed convolution is to apply an
equivalent -- but much less efficient -- direct convolution. The example
described so far could be tackled by convolving a $3 \times 3$ kernel over a
$2 \times 2$ input padded with a $2 \times 2$ border of zeros using unit
strides (i.e., $i' = 2$, $k' = k$, $s' = 1$ and $p' = 2$), as shown in
\autoref{fig:no_padding_no_strides_transposed}. Notably, the kernel's and
stride's sizes remain the same, but the input of the transposed convolution is
now zero padded.\footnote{Note that although
    equivalent to applying the transposed matrix, this visualization adds a lot
    of zero multiplications in the form of zero padding.  This is done here for
    illustration purposes, but it is inefficient, and software implementations
    will normally not perform the useless zero multiplications.}

One way to understand the logic behind zero padding is to consider the
connectivity pattern of the transposed convolution and use it to guide the
design of the equivalent convolution. For example, the top left pixel of the
input of the direct convolution only contribute to the top left pixel of the
output, the top right pixel is only connected to the top right output pixel,
and so on.

To maintain the same connectivity pattern in the equivalent convolution it is
necessary to zero pad the input in such a way that the first (top-left)
application of the kernel only touches the top-left pixel, i.e., the padding
has to be equal to the size of the kernel minus one.

Proceeding in the same fashion it is possible to determine similar observations
for the other elements of the image, giving rise to the following relationship:

\begin{relationship}\label{rel:no_padding_no_strides_transposed}
A convolution described by $s = 1$, $p = 0$ and $k$ has an associated
transposed convolution described by $k' = k$, $s' = s$ and $p' = k - 1$ and its
output size is
\begin{equation*}
    o' = i' + (k - 1).
\end{equation*}
\end{relationship}

Interestingly, this corresponds to a fully padded convolution with unit
strides.

\section{Zero padding, unit strides, transposed}

Knowing that the transpose of a non-padded convolution is equivalent to
convolving a zero padded input, it would be reasonable to suppose that the
transpose of a zero padded convolution is equivalent to convolving an input
padded with {\em less\/} zeros.

\begin{figure}[p]
    \centering
    \includegraphics[width=0.24\textwidth]{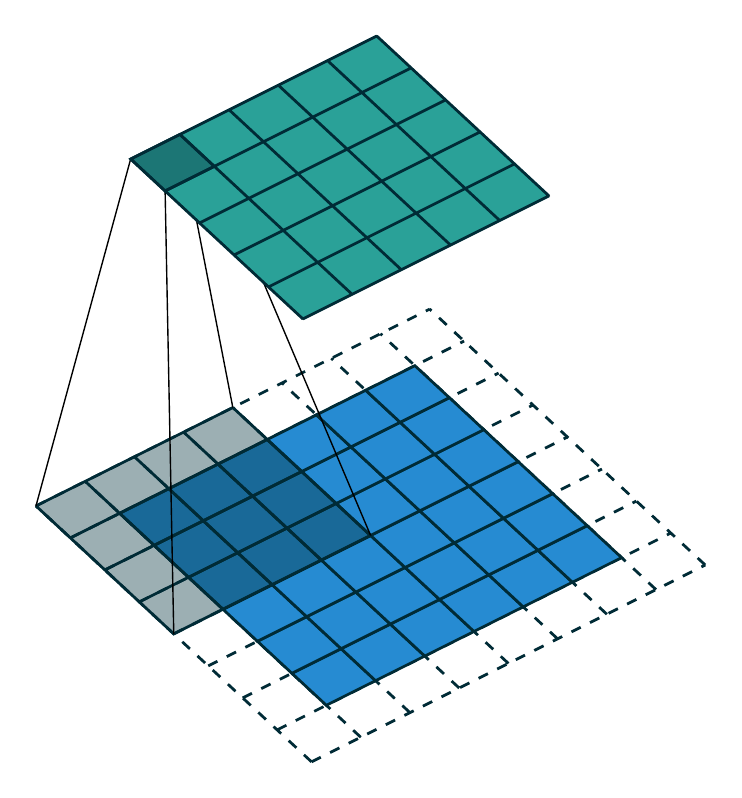}
    \includegraphics[width=0.24\textwidth]{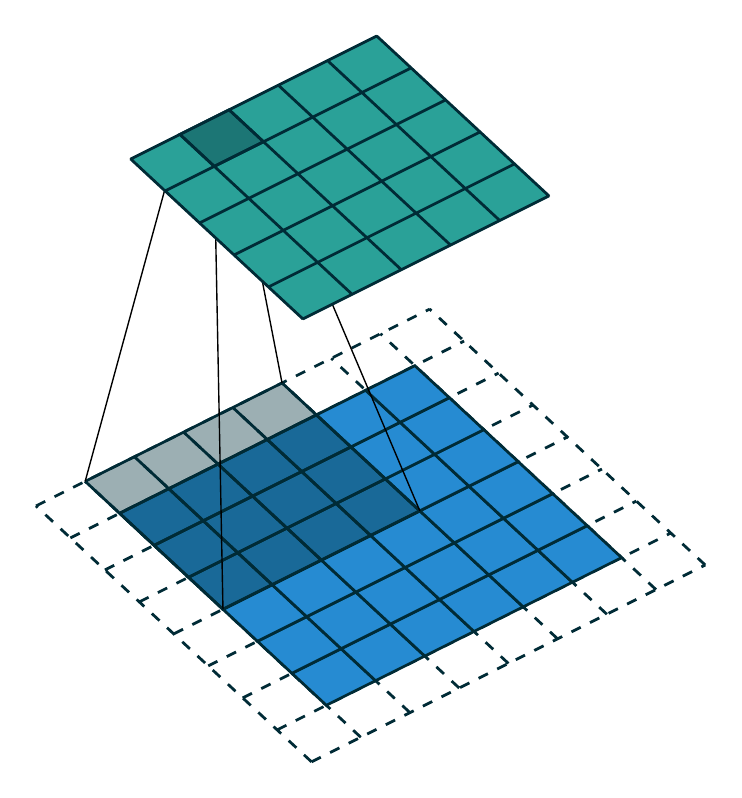}
    \includegraphics[width=0.24\textwidth]{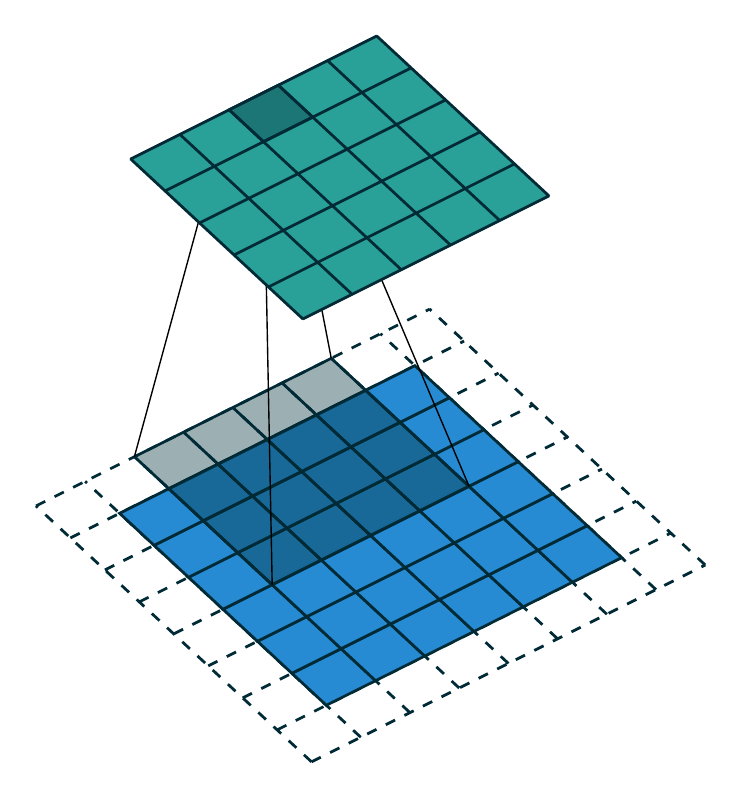}
    \includegraphics[width=0.24\textwidth]{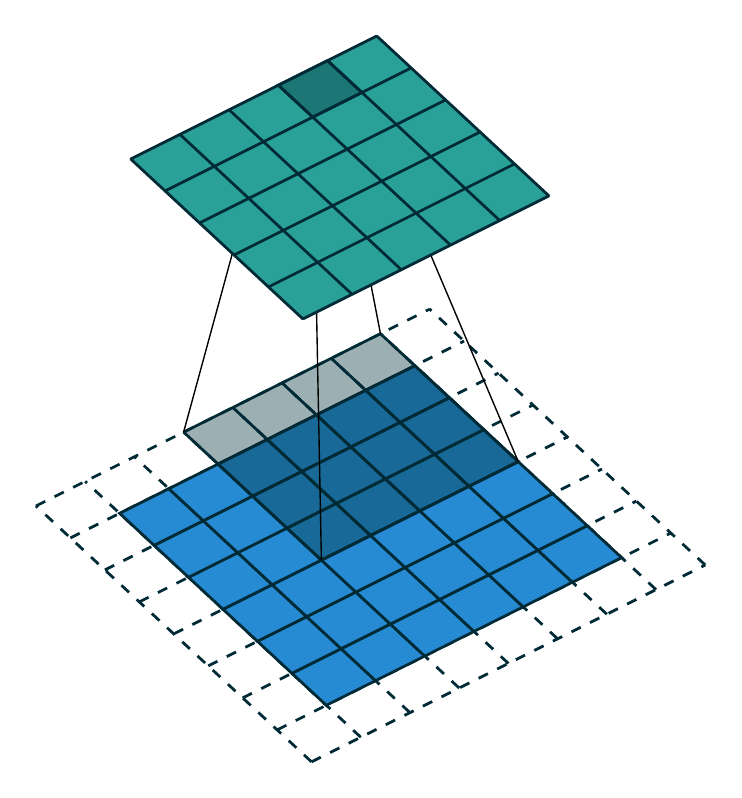}
    \caption{\label{fig:arbitrary_padding_no_strides_transposed} The transpose
        of convolving a $4 \times 4$ kernel over a $5 \times 5$ input padded
        with a $2 \times 2$ border of zeros using unit strides (i.e., $i = 5$,
        $k = 4$, $s = 1$ and $p = 2$). It is equivalent to convolving a $4
        \times 4$ kernel over a $6 \times 6$ input padded with a $1 \times 1$
        border of zeros using unit strides (i.e., $i' = 6$, $k' = k$, $s' = 1$
        and $p' = 1$).}
\end{figure}

It is indeed the case, as shown in
\autoref{fig:arbitrary_padding_no_strides_transposed} for $i = 5$, $k = 4$ and
$p = 2$.

Formally, the following relationship applies for zero padded convolutions:

\begin{relationship}\label{rel:arbitrary_padding_no_strides_transposed}
A convolution described by $s = 1$, $k$ and $p$ has an
associated transposed convolution described by $k' = k$, $s' = s$ and $p' = k -
p - 1$ and its output size is
\begin{equation*}
    o' = i' + (k - 1) - 2p.
\end{equation*}
\end{relationship}

\begin{figure}[p]
    \centering
    \includegraphics[width=0.24\textwidth]{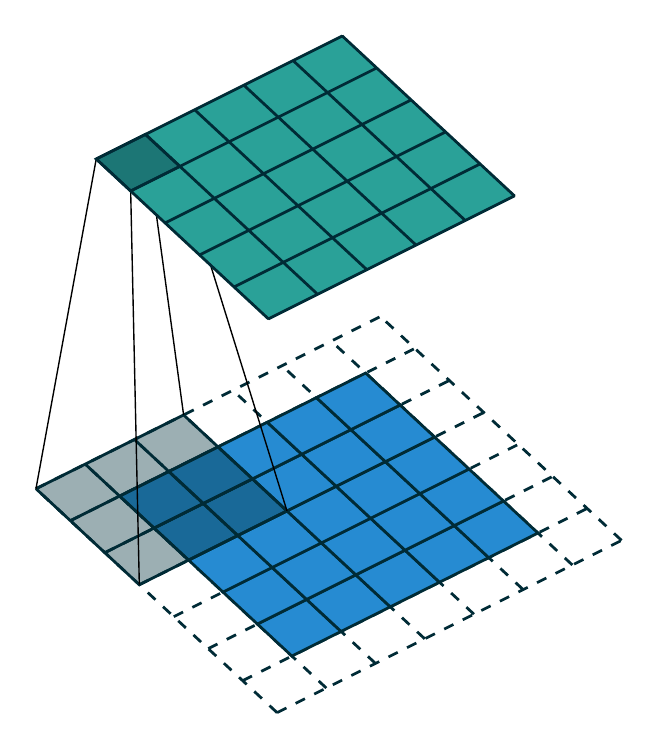}
    \includegraphics[width=0.24\textwidth]{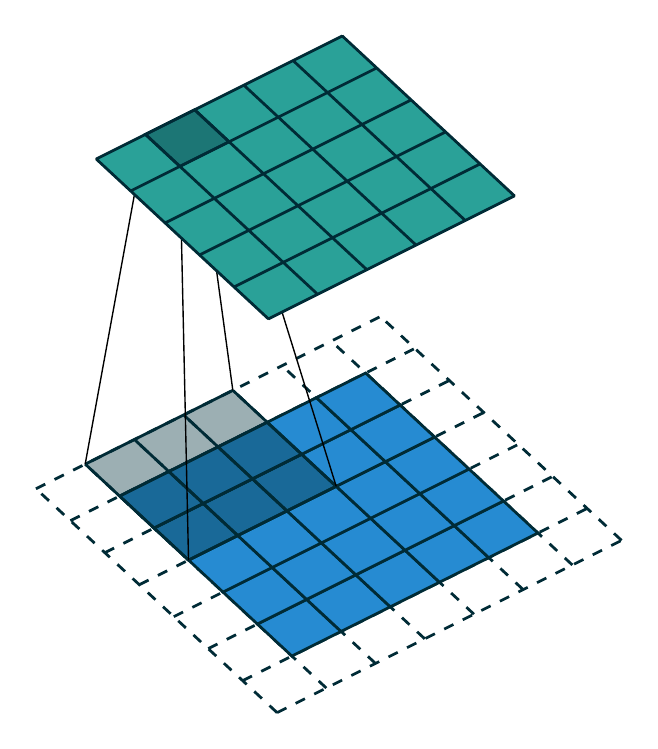}
    \includegraphics[width=0.24\textwidth]{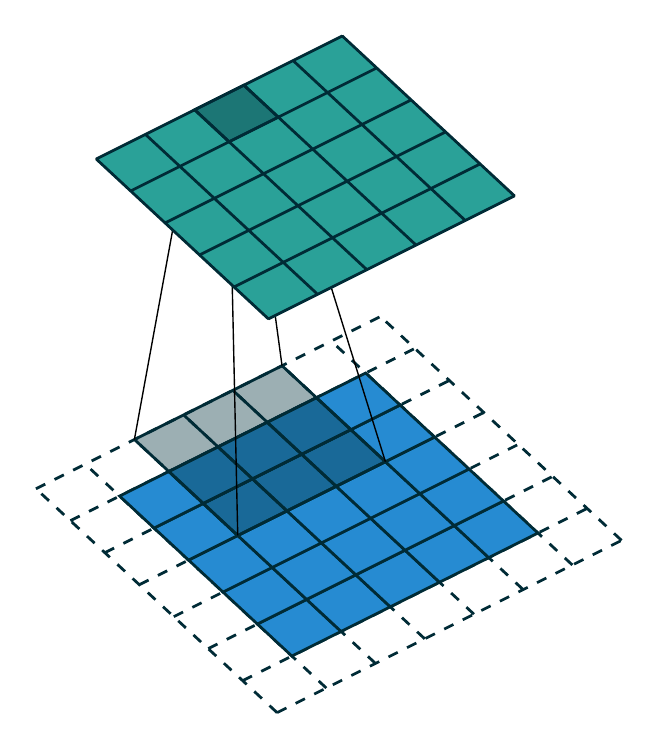}
    \includegraphics[width=0.24\textwidth]{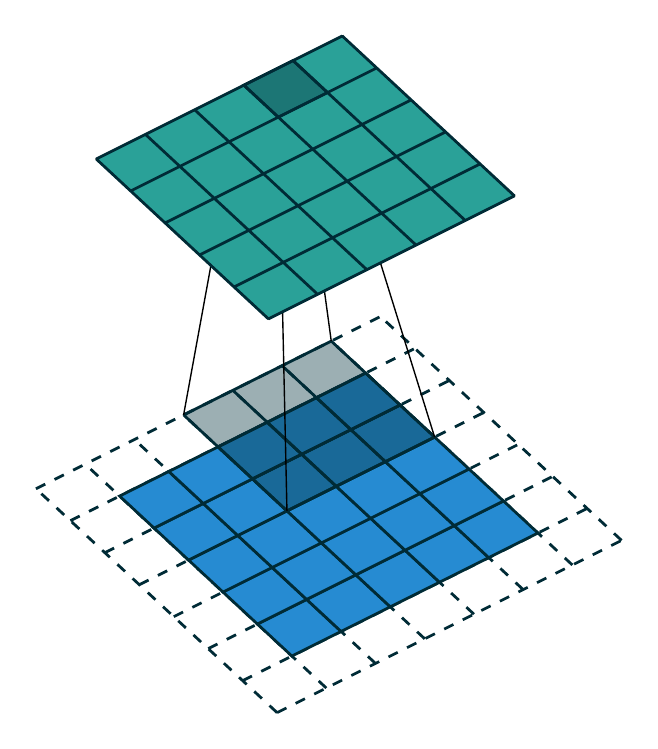}
    \caption{\label{fig:same_padding_no_strides_transposed} The transpose of
        convolving a $3 \times 3$ kernel over a $5 \times 5$ input using half
        padding and unit strides (i.e., $i = 5$, $k = 3$, $s = 1$ and $p = 1$).
        It is equivalent to convolving a $3 \times 3$ kernel over a $5 \times 5$
        input using half padding and unit strides (i.e., $i' = 5$, $k' = k$, $s'
        = 1$ and $p' = 1$).}
\end{figure}

\subsection{Half (same) padding, transposed}

By applying the same inductive reasoning as before, it is reasonable to expect
that the equivalent convolution of the transpose of a half padded convolution
is itself a half padded convolution, given that the output size of a half
padded convolution is the same as its input size. Thus the following relation
applies:

\begin{relationship}\label{rel:half_padding_no_strides_transposed}
A convolution described by $k = 2n + 1, \quad n \in \mathbb{N}$, $s = 1$ and $p
= \lfloor k / 2 \rfloor = n$ has an associated transposed convolution described
by $k' = k$, $s' = s$ and $p' = p$ and its output size is
\begin{equation*}
\begin{split}
    o' &= i' + (k - 1) - 2p \\
       &= i' + 2n - 2n \\
       &= i'.
\end{split}
\end{equation*}
\end{relationship}

\autoref{fig:same_padding_no_strides_transposed} provides an example for $i =
5$, $k = 3$ and (therefore) $p = 1$.

\subsection{Full padding, transposed}

Knowing that the equivalent convolution of the transpose of a non-padded
convolution involves full padding, it is unsurprising that the equivalent of
the transpose of a fully padded convolution is a non-padded convolution:

\begin{relationship}\label{rel:full_padding_no_strides_transposed}
A convolution described by $s = 1$, $k$ and $p = k - 1$ has an
associated transposed convolution described by $k' = k$, $s' = s$ and $p' = 0$
and its output size is
\begin{equation*}
\begin{split}
    o' &= i' + (k - 1) - 2p \\
       &= i' - (k - 1)
\end{split}
\end{equation*}
\end{relationship}

\begin{figure}[p]
    \centering
    \includegraphics[width=0.24\textwidth]{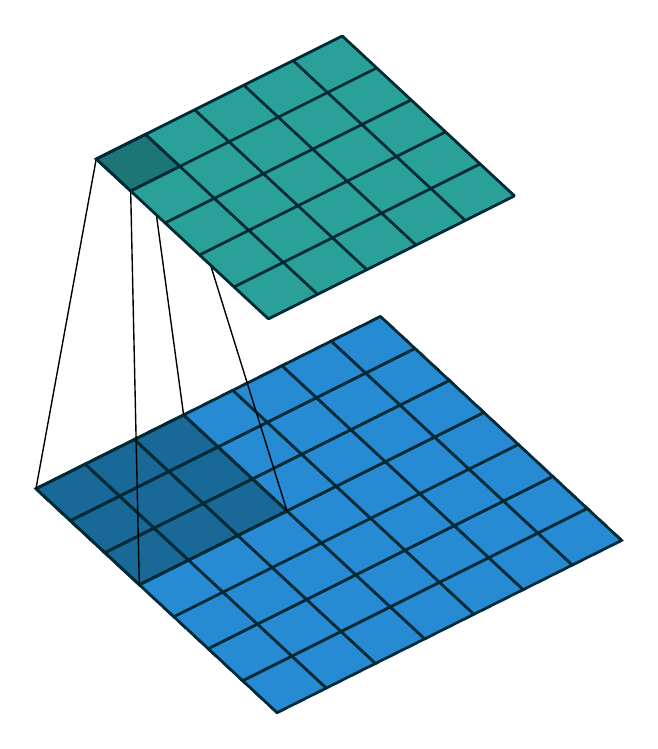}
    \includegraphics[width=0.24\textwidth]{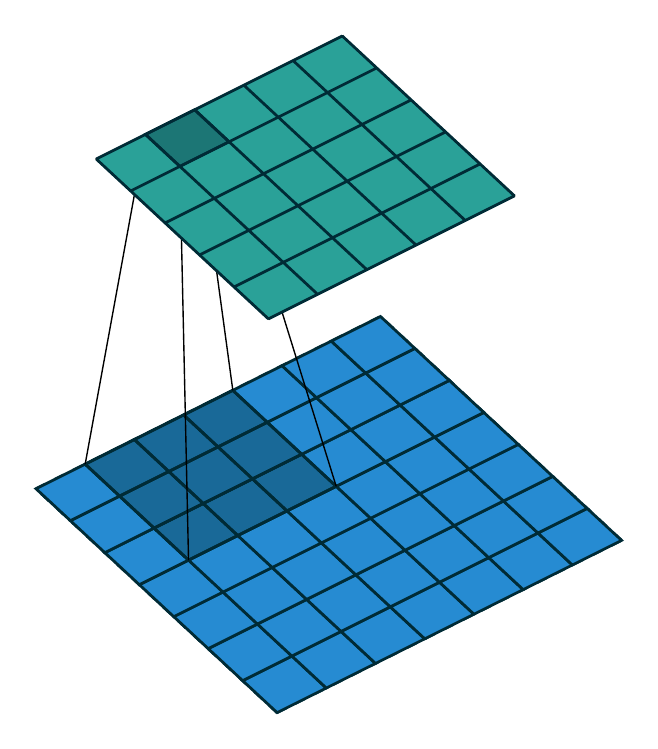}
    \includegraphics[width=0.24\textwidth]{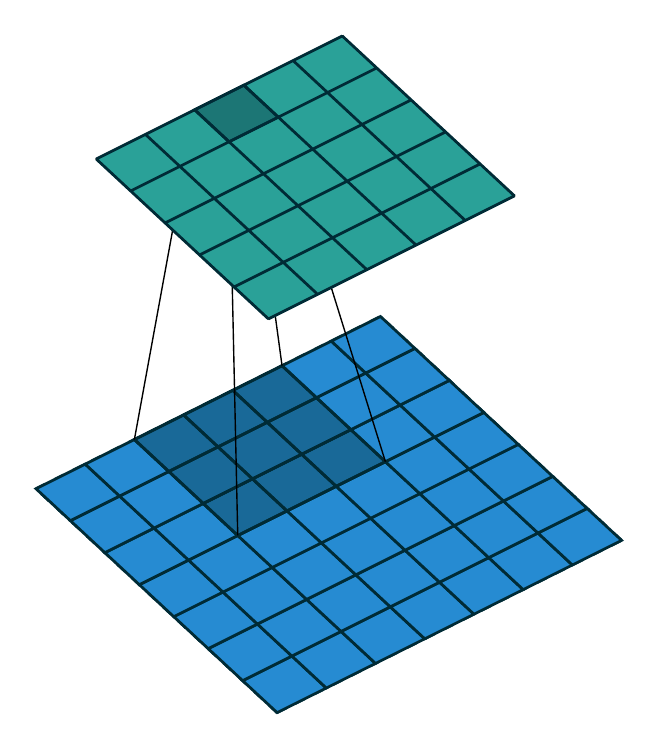}
    \includegraphics[width=0.24\textwidth]{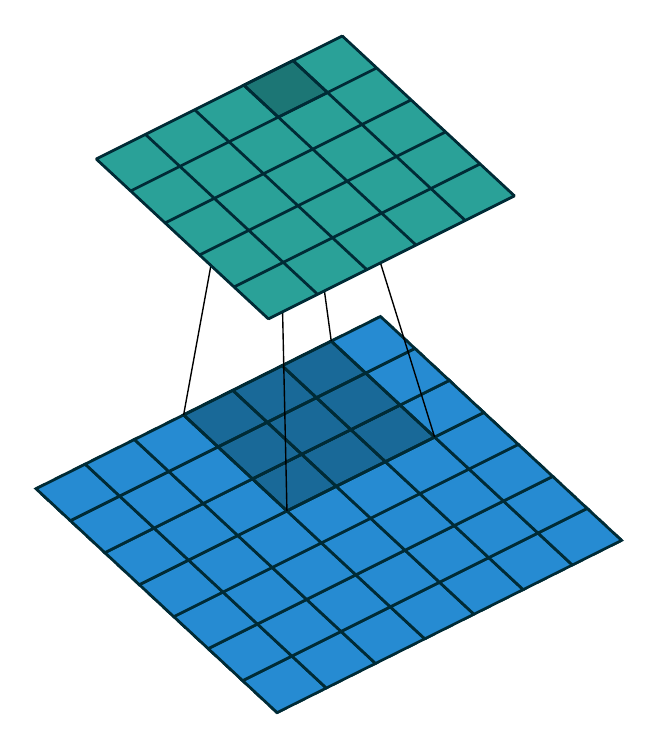}
    \caption{\label{fig:full_padding_no_strides_transposed} The transpose of
        convolving a $3 \times 3$ kernel over a $5 \times 5$ input using full
        padding and unit strides (i.e., $i = 5$, $k = 3$, $s = 1$ and $p = 2$).
        It is equivalent to convolving a $3 \times 3$ kernel over a $7 \times 7$
        input using unit strides (i.e., $i' = 7$, $k' = k$, $s' = 1$ and $p' =
        0$).}
\end{figure}

\autoref{fig:full_padding_no_strides_transposed} provides an example for $i =
5$, $k = 3$ and (therefore) $p = 2$.

\section{No zero padding, non-unit strides, transposed}

Using the same kind of inductive logic as for zero padded convolutions, one
might expect that the transpose of a convolution with $s > 1$ involves an
equivalent convolution with $s < 1$. As will be explained, this is a valid
intuition, which is why transposed convolutions are sometimes called {\em
fractionally strided convolutions}.

\begin{figure}[p]
    \centering
    \includegraphics[width=0.24\textwidth]{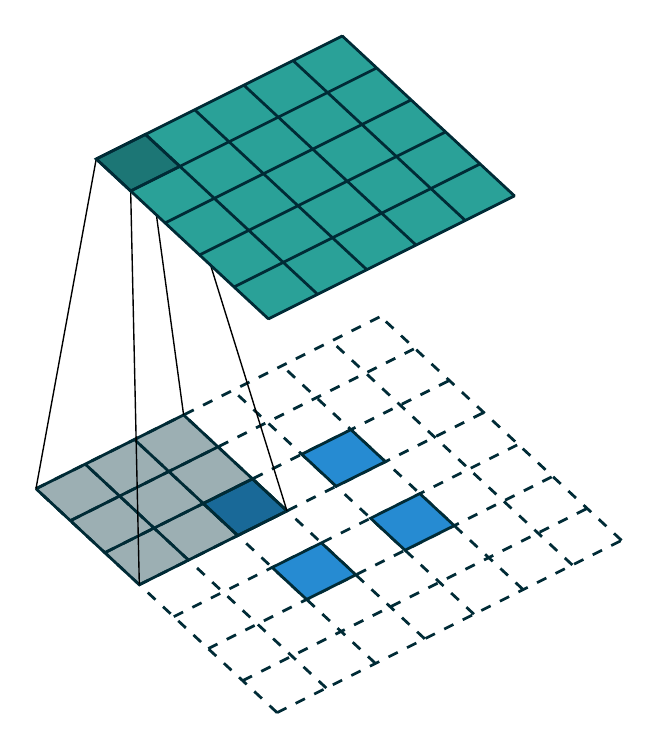}
    \includegraphics[width=0.24\textwidth]{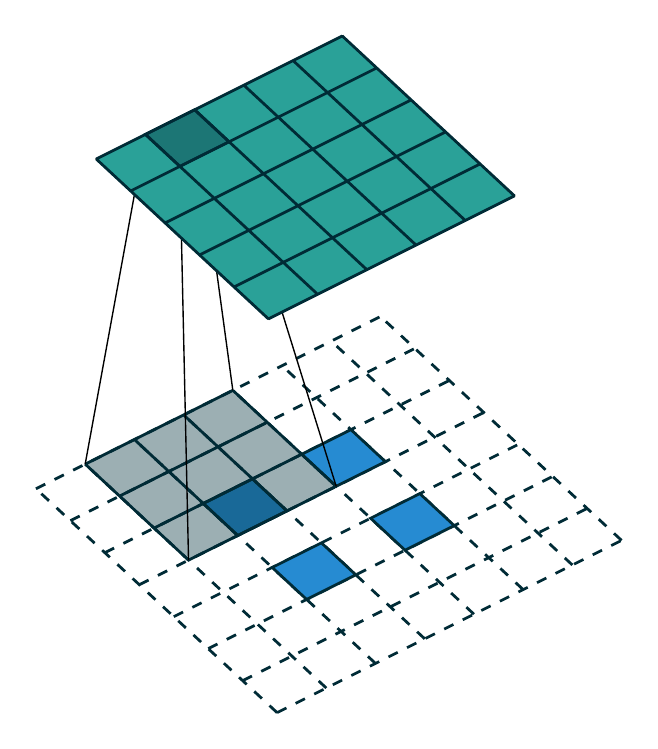}
    \includegraphics[width=0.24\textwidth]{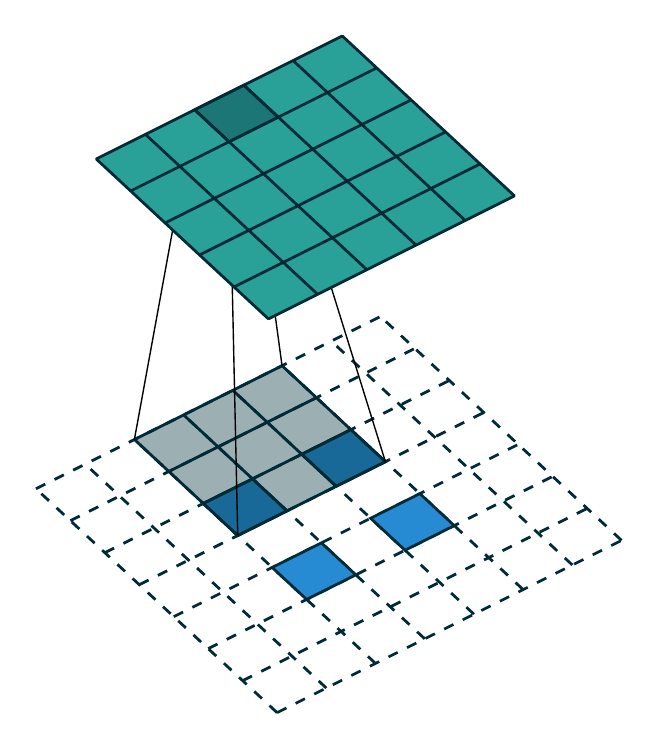}
    \includegraphics[width=0.24\textwidth]{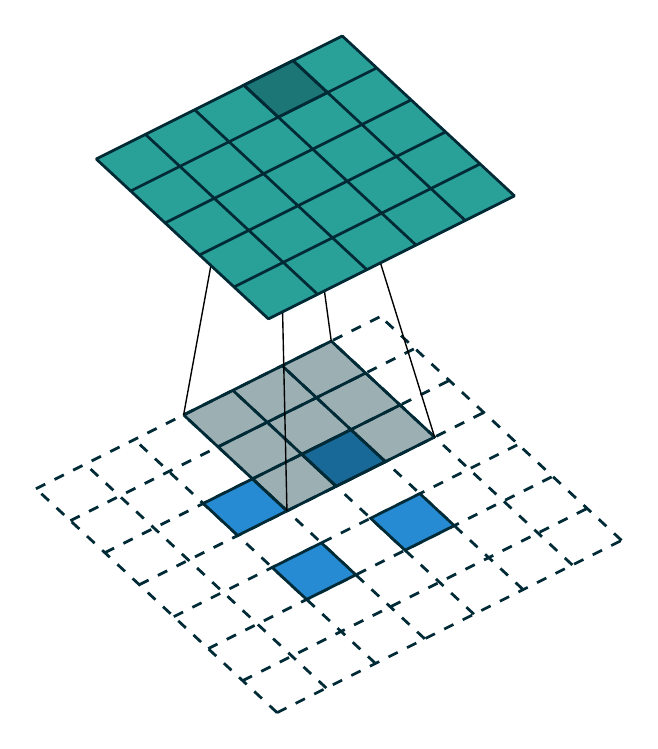}
    \caption{\label{fig:no_padding_strides_transposed} The transpose of
        convolving a $3 \times 3$ kernel over a $5 \times 5$ input using $2
        \times 2$ strides (i.e., $i = 5$, $k = 3$, $s = 2$ and $p = 0$). It is
        equivalent to convolving a $3 \times 3$ kernel over a $2 \times 2$ input
        (with $1$ zero inserted between inputs) padded with a $2 \times 2$
        border of zeros using unit strides (i.e., $i' = 2$, $\tilde{i}' = 3$, $k'
        = k$, $s' = 1$ and $p' = 2$).}
\end{figure}

\autoref{fig:no_padding_strides_transposed} provides an example for $i = 5$, $k
= 3$ and $s = 2$ which helps understand what fractional strides involve: zeros
are inserted {\em between\/} input units, which makes the kernel move around at
a slower pace than with unit strides.\footnote{Doing so is inefficient and
    real-world implementations avoid useless multiplications by zero, but
    conceptually it is how the transpose of a strided convolution can be
    thought of.}

For the moment, it will be assumed that the convolution is non-padded ($p = 0$)
and that its input size $i$ is such that $i - k$ is a multiple of $s$. In that
case, the following relationship holds:

\begin{relationship}\label{rel:no_padding_strides_transposed}
A convolution described by $p = 0$, $k$ and $s$ and whose input
size is such that $i - k$ is a multiple of $s$, has an associated transposed
convolution described by $\tilde{i}'$, $k' = k$, $s' = 1$ and $p' = k - 1$,
where $\tilde{i}'$ is the size of the stretched input obtained by adding
$s - 1$ zeros between each input unit, and its output size is
\begin{equation*}
\begin{split}
    o' = s (i' - 1) + k.
\end{split}
\end{equation*}
\end{relationship}

\begin{figure}[p]
    \centering
    \includegraphics[width=0.24\textwidth]{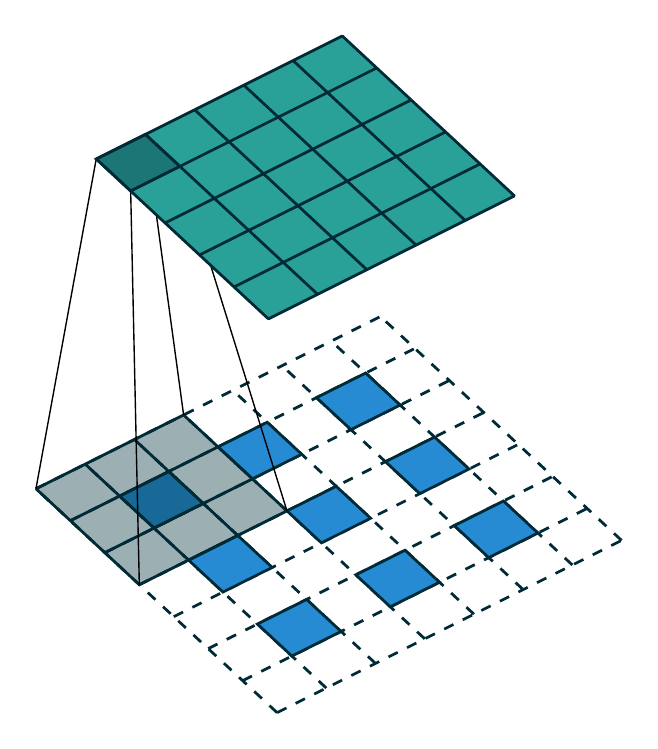}
    \includegraphics[width=0.24\textwidth]{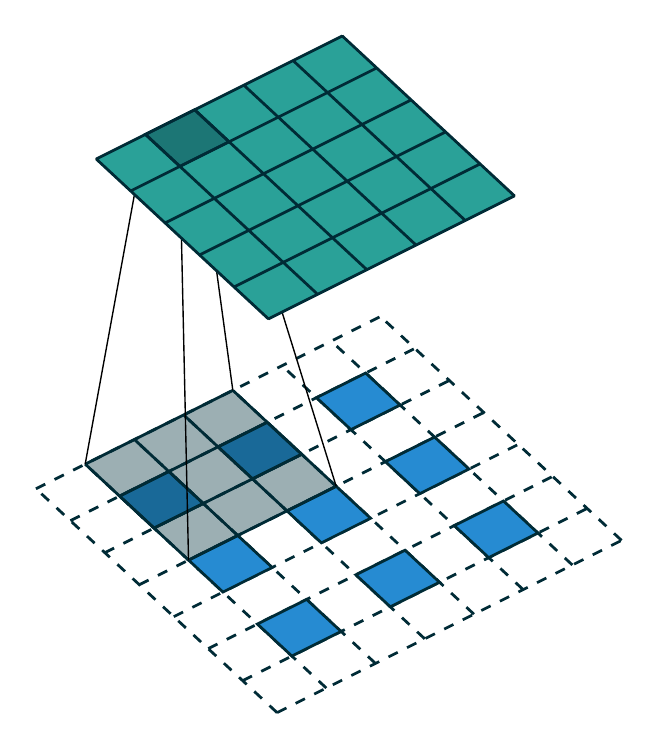}
    \includegraphics[width=0.24\textwidth]{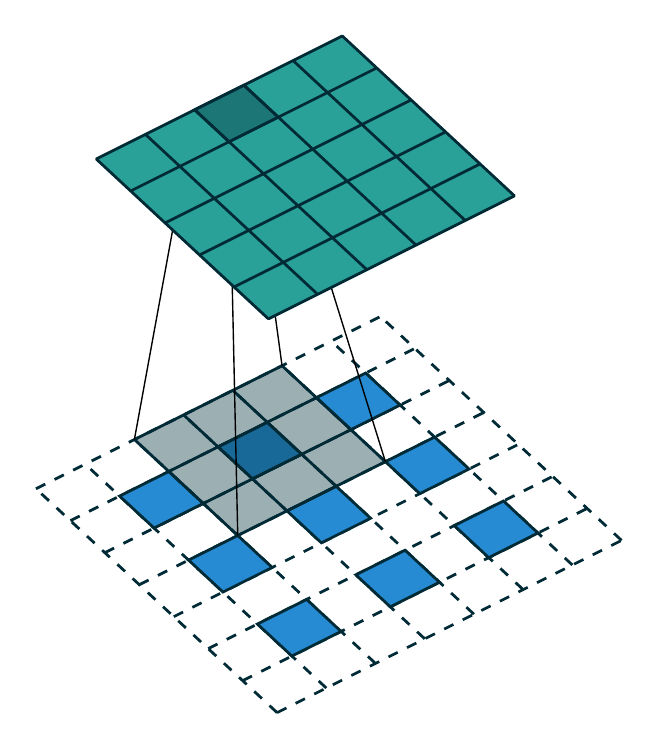}
    \includegraphics[width=0.24\textwidth]{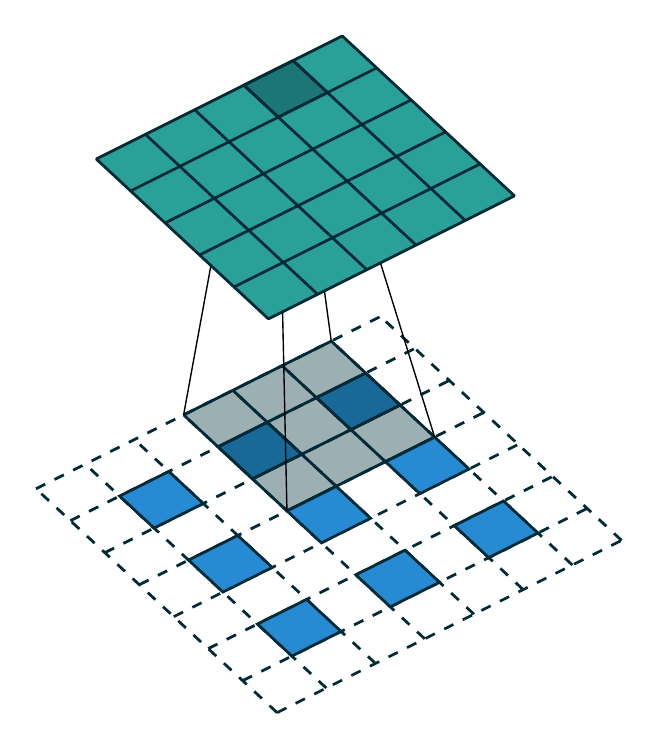}
    \caption{\label{fig:padding_strides_transposed} The transpose of convolving
        a $3 \times 3$ kernel over a $5 \times 5$ input padded with a $1 \times
        1$ border of zeros using $2 \times 2$ strides (i.e., $i = 5$, $k = 3$, $s
        = 2$ and $p = 1$). It is equivalent to convolving a $3 \times 3$ kernel
        over a $3 \times 3$ input (with $1$ zero inserted between inputs) padded
        with a $1 \times 1$ border of zeros using unit strides (i.e., $i' = 3$,
        $\tilde{i}' = 5$, $k' = k$, $s' = 1$ and $p' = 1$).}
\end{figure}

\section{Zero padding, non-unit strides, transposed}

When the convolution's input size $i$ is such that $i + 2p - k$ is a multiple
of $s$, the analysis can extended to the zero padded case by combining
\autoref{rel:arbitrary_padding_no_strides_transposed} and
\autoref{rel:no_padding_strides_transposed}:

\begin{relationship}\label{rel:padding_strides_transposed}
A convolution described by $k$, $s$ and $p$ and whose
input size $i$ is such that $i + 2p - k$ is a multiple of $s$ has an associated
transposed convolution described by $\tilde{i}'$, $k' = k$, $s' = 1$ and
$p' = k - p - 1$, where $\tilde{i}'$ is the size of the stretched input
obtained by adding $s - 1$ zeros between each input unit, and its output size
is
\begin{equation*}
\begin{split}
    o' = s (i' - 1) + k - 2p.
\end{split}
\end{equation*}
\end{relationship}

\autoref{fig:padding_strides_transposed} provides an example for $i = 5$, $k =
3$, $s = 2$ and $p = 1$.

The constraint on the size of the input $i$ can be relaxed by introducing
another parameter $a \in \{0, \ldots, s - 1\}$ that allows to distinguish
between the $s$ different cases that all lead to the same $i'$:

\begin{relationship}\label{rel:padding_strides_transposed_odd}
A convolution described by $k$, $s$ and $p$ has an
associated transposed convolution described by $a$, $\tilde{i}'$, $k' = k$, $s'
= 1$ and $p' = k - p - 1$, where $\tilde{i}'$ is the size of the stretched
input obtained by adding $s - 1$ zeros between each input unit, and $a = (i +
2p - k) \mod s$ represents the number of zeros added to the bottom and right edges
of the input, and its output size is
\begin{equation*}
\begin{split}
    o' = s (i' - 1) + a + k - 2p.
\end{split}
\end{equation*}
\end{relationship}

\begin{figure}[p]
    \centering
    \includegraphics[width=0.24\textwidth]{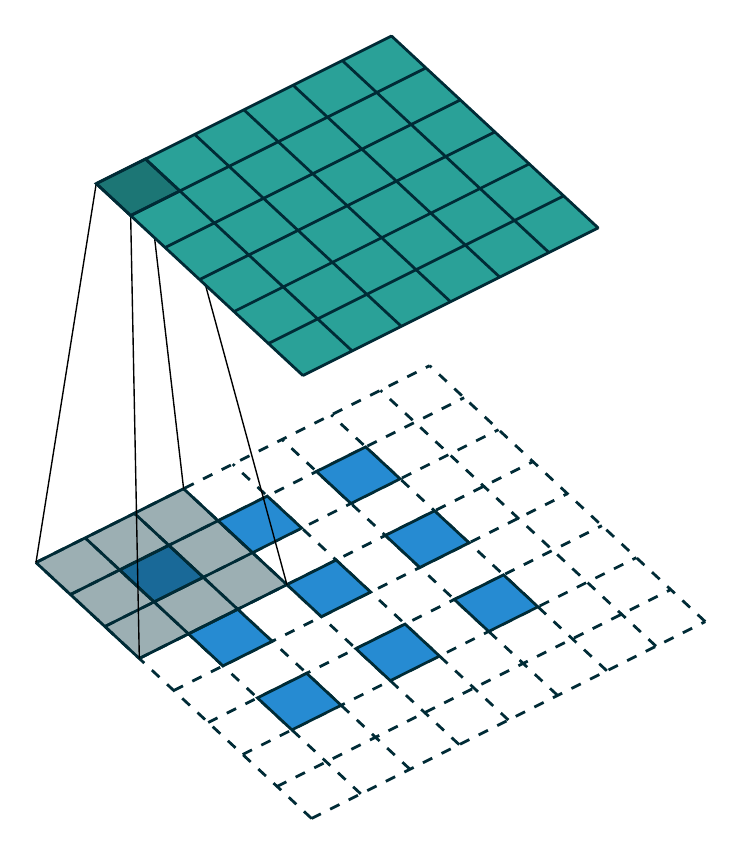}
    \includegraphics[width=0.24\textwidth]{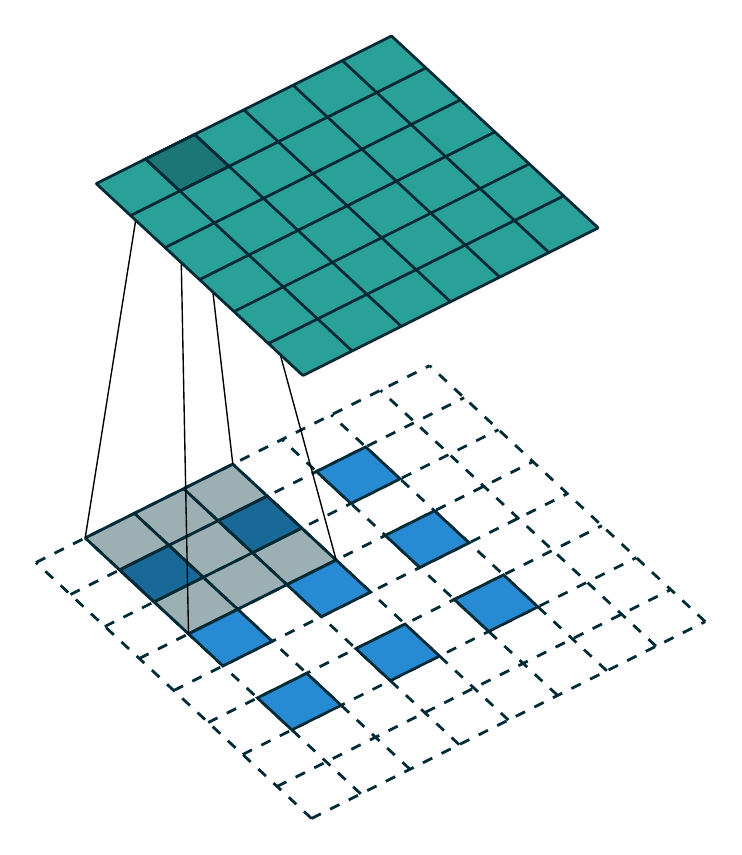}
    \includegraphics[width=0.24\textwidth]{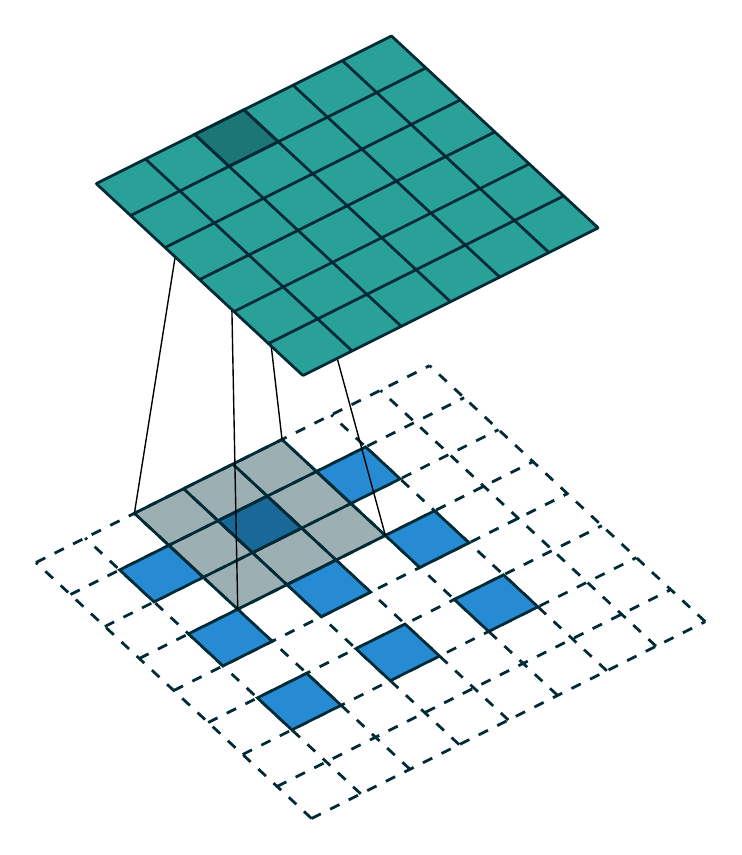}
    \includegraphics[width=0.24\textwidth]{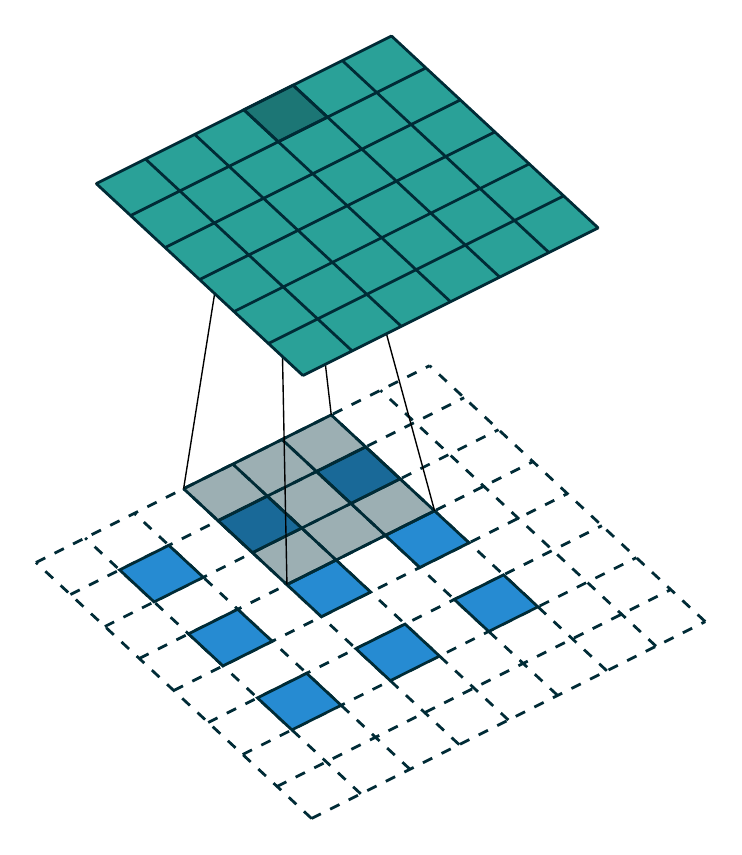}
    \caption{\label{fig:padding_strides_odd_transposed} The transpose of
        convolving a $3 \times 3$ kernel over a $6 \times 6$ input padded with a
        $1 \times 1$ border of zeros using $2 \times 2$ strides (i.e., $i = 6$,
        $k = 3$, $s = 2$ and $p = 1$). It is equivalent to convolving a $3
        \times 3$ kernel over a $2 \times 2$ input (with $1$ zero inserted
        between inputs) padded with a $1 \times 1$ border of zeros (with an
        additional border of size $1$ added to the bottom and right edges) using
        unit strides (i.e., $i' = 3$, $\tilde{i}' = 5$, $a = 1$, $k' = k$, $s' =
        1$ and $p' = 1$).}
\end{figure}

\autoref{fig:padding_strides_odd_transposed} provides an example for $i = 6$, $k
= 3$, $s = 2$ and $p = 1$.

\chapter{Miscellaneous convolutions}

\section{Dilated convolutions}

Readers familiar with the deep learning literature may have noticed the term
``dilated convolutions'' (or ``atrous convolutions'', from the French expression
{\em convolutions \`{a} trous}) appear in recent papers. Here we attempt to
provide an intuitive understanding of dilated convolutions. For a more in-depth
description and to understand in what contexts they are applied, see
\citet{chen2014semantic,yu2015multi}.

Dilated convolutions ``inflate'' the kernel by inserting spaces between the
kernel elements. The dilation ``rate'' is controlled by an additional
hyperparameter $d$. Implementations may vary, but there are usually $d - 1$
spaces inserted between kernel elements such that $d = 1$ corresponds to a
regular convolution.

Dilated convolutions are used to cheaply increase the receptive field of output
units without increasing the kernel size, which is especially effective
when multiple dilated convolutions are stacked one after another. For a
concrete example, see \citet{oord2016wavenet}, in which the proposed WaveNet
model implements an autoregressive generative model for raw audio which uses
dilated convolutions to condition new audio frames on a large context of past
audio frames.

To understand the relationship tying the dilation rate $d$ and the output size
$o$, it is useful to think of the impact of $d$ on the {\em effective kernel
size}. A kernel of size $k$ dilated by a factor $d$ has an effective size
\begin{equation*}
    \hat{k} = k + (k - 1)(d - 1).
\end{equation*}
This can be combined with \autoref{rel:padding_strides} to form the following
relationship for dilated convolutions:

\begin{relationship}\label{rel:dilation}
For any $i$, $k$, $p$ and $s$, and for a dilation rate $d$,
\begin{equation*}
    o = \left\lfloor \frac{i + 2p - k - (k - 1)(d - 1)}{s} \right\rfloor + 1.
\end{equation*}
\end{relationship}

\begin{figure}[h]
    \centering
    \includegraphics[width=0.24\textwidth]{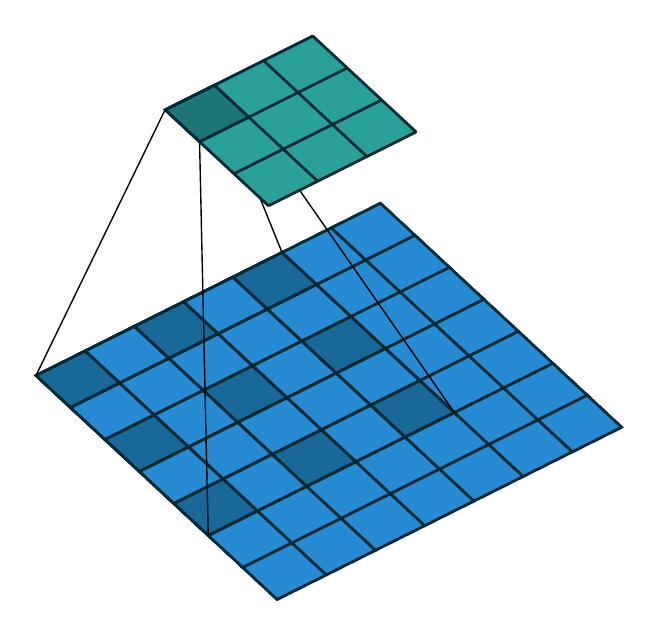}
    \includegraphics[width=0.24\textwidth]{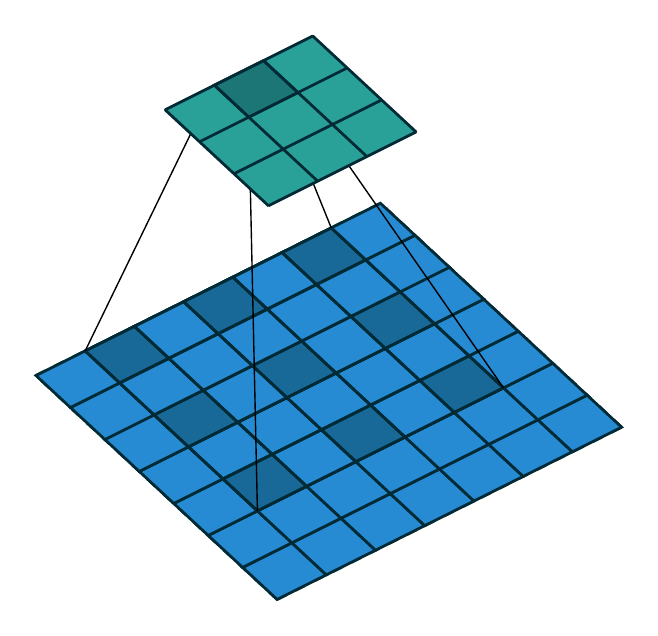}
    \includegraphics[width=0.24\textwidth]{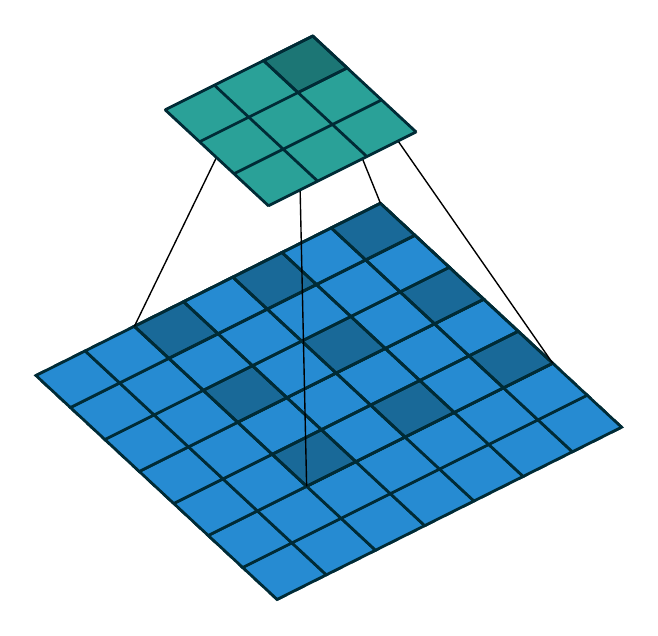}
    \includegraphics[width=0.24\textwidth]{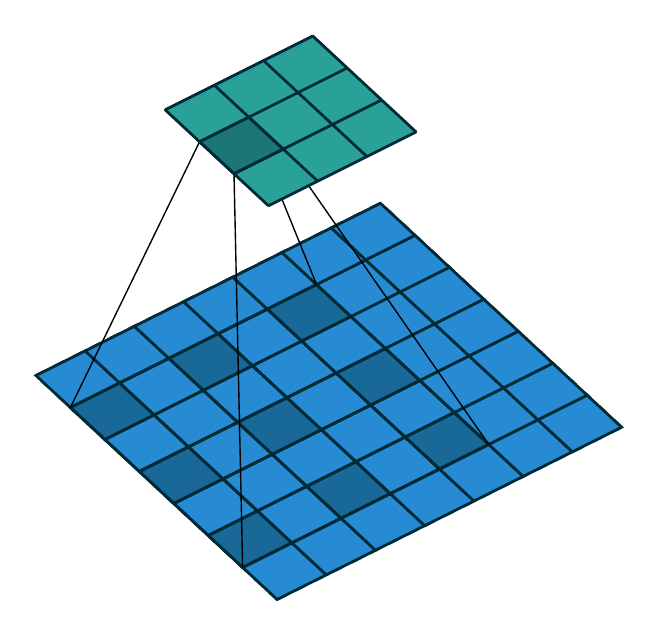}
    \caption{\label{fig:dilation} (Dilated convolution)
        Convolving a $3 \times 3$ kernel over a $7 \times 7$ input with a
        dilation factor of 2 (i.e., $i = 7$, $k = 3$, $d = 2$, $s = 1$ and
        $p = 0$).}
\end{figure}

\noindent \autoref{fig:dilation} provides an example for $i = 7$, $k = 3$ and
$d = 2$.

\bibliography{bibliography}

\begin{thebibliography}{}

\bibitem[Abadi {\em et~al.}(2015)Abadi, Agarwal, Barham, Brevdo, Chen, Citro,
  Corrado, Davis, Dean, Devin, {\em et~al.}]{abaditensorflow}
Abadi, M., Agarwal, A., Barham, P., Brevdo, E., Chen, Z., Citro, C., Corrado,
  G.~S., Davis, A., Dean, J., Devin, M., {\em et~al.} (2015).
\newblock Tensorflow: Large-scale machine learning on heterogeneous systems.
\newblock {\em Software available from tensorflow.org\/}.

\bibitem[Bastien {\em et~al.}(2012)Bastien, Lamblin, Pascanu, Bergstra,
  Goodfellow, Bergeron, Bouchard, Warde-Farley, and Bengio]{bastien2012theano}
Bastien, F., Lamblin, P., Pascanu, R., Bergstra, J., Goodfellow, I., Bergeron,
  A., Bouchard, N., Warde-Farley, D., and Bengio, Y. (2012).
\newblock Theano: new features and speed improvements.
\newblock {\em arXiv preprint arXiv:1211.5590\/}.

\bibitem[Bergstra {\em et~al.}(2010)Bergstra, Breuleux, Bastien, Lamblin,
  Pascanu, Desjardins, Turian, Warde-Farley, and Bengio]{bergstra2010theano}
Bergstra, J., Breuleux, O., Bastien, F., Lamblin, P., Pascanu, R., Desjardins,
  G., Turian, J., Warde-Farley, D., and Bengio, Y. (2010).
\newblock Theano: A cpu and gpu math compiler in python.
\newblock In {\em Proc. 9th Python in Science Conf\/}, pages 1--7.

\bibitem[Boureau {\em et~al.}(2010a)Boureau, Bach, LeCun, and
  Ponce]{boureau-cvpr-10}
Boureau, Y., Bach, F., LeCun, Y., and Ponce, J. (2010a).
\newblock Learning mid-level features for recognition.
\newblock In {\em Proc. International Conference on Computer Vision and Pattern
  Recognition (CVPR'10)\/}. IEEE.

\bibitem[Boureau {\em et~al.}(2010b)Boureau, Ponce, and LeCun]{boureau-icml-10}
Boureau, Y., Ponce, J., and LeCun, Y. (2010b).
\newblock A theoretical analysis of feature pooling in vision algorithms.
\newblock In {\em Proc. International Conference on Machine learning
  (ICML'10)\/}.

\bibitem[Boureau {\em et~al.}(2011)Boureau, {Le Roux}, Bach, Ponce, and
  LeCun]{boureau-iccv-11}
Boureau, Y., {Le Roux}, N., Bach, F., Ponce, J., and LeCun, Y. (2011).
\newblock Ask the locals: multi-way local pooling for image recognition.
\newblock In {\em Proc. International Conference on Computer Vision
  (ICCV'11)\/}. IEEE.

\bibitem[Chen {\em et~al.}(2014)Chen, Papandreou, Kokkinos, Murphy, and
  Yuille]{chen2014semantic}
Chen, L.-C., Papandreou, G., Kokkinos, I., Murphy, K., and Yuille, A.~L.
  (2014).
\newblock Semantic image segmentation with deep convolutional nets and fully
  connected crfs.
\newblock {\em arXiv preprint arXiv:1412.7062\/}.

\bibitem[Collobert {\em et~al.}(2011)Collobert, Kavukcuoglu, and
  Farabet]{collobert2011torch7}
Collobert, R., Kavukcuoglu, K., and Farabet, C. (2011).
\newblock Torch7: A matlab-like environment for machine learning.
\newblock In {\em BigLearn, NIPS Workshop\/}, number EPFL-CONF-192376.

\bibitem[Goodfellow {\em et~al.}(2016)Goodfellow, Bengio, and
  Courville]{Goodfellow-et-al-2016-Book}
Goodfellow, I., Bengio, Y., and Courville, A. (2016).
\newblock Deep learning.
\newblock Book in preparation for MIT Press.

\bibitem[Im {\em et~al.}(2016)Im, Kim, Jiang, and Memisevic]{im2016generating}
Im, D.~J., Kim, C.~D., Jiang, H., and Memisevic, R. (2016).
\newblock Generating images with recurrent adversarial networks.
\newblock {\em arXiv preprint arXiv:1602.05110\/}.

\bibitem[Jia {\em et~al.}(2014)Jia, Shelhamer, Donahue, Karayev, Long,
  Girshick, Guadarrama, and Darrell]{jia2014caffe}
Jia, Y., Shelhamer, E., Donahue, J., Karayev, S., Long, J., Girshick, R.,
  Guadarrama, S., and Darrell, T. (2014).
\newblock Caffe: Convolutional architecture for fast feature embedding.
\newblock In {\em Proceedings of the ACM International Conference on
  Multimedia\/}, pages 675--678. ACM.

\bibitem[Krizhevsky {\em et~al.}(2012)Krizhevsky, Sutskever, and
  Hinton]{krizhevsky2012imagenet}
Krizhevsky, A., Sutskever, I., and Hinton, G.~E. (2012).
\newblock Imagenet classification with deep convolutional neural networks.
\newblock In {\em Advances in neural information processing systems\/}, pages
  1097--1105.

\bibitem[Le~Cun {\em et~al.}(1997)Le~Cun, Bottou, and Bengio]{le1997reading}
Le~Cun, Y., Bottou, L., and Bengio, Y. (1997).
\newblock Reading checks with multilayer graph transformer networks.
\newblock In {\em Acoustics, Speech, and Signal Processing, 1997. ICASSP-97.,
  1997 IEEE International Conference on\/}, volume~1, pages 151--154. IEEE.

\bibitem[Long {\em et~al.}(2015)Long, Shelhamer, and Darrell]{long2015fully}
Long, J., Shelhamer, E., and Darrell, T. (2015).
\newblock Fully convolutional networks for semantic segmentation.
\newblock In {\em Proceedings of the IEEE Conference on Computer Vision and
  Pattern Recognition\/}, pages 3431--3440.

\bibitem[Oord {\em et~al.}(2016)Oord, Dieleman, Zen, Simonyan, Vinyals, Graves,
  Kalchbrenner, Senior, and Kavukcuoglu]{oord2016wavenet}
Oord, A. v.~d., Dieleman, S., Zen, H., Simonyan, K., Vinyals, O., Graves, A.,
  Kalchbrenner, N., Senior, A., and Kavukcuoglu, K. (2016).
\newblock Wavenet: A generative model for raw audio.
\newblock {\em arXiv preprint arXiv:1609.03499\/}.

\bibitem[Radford {\em et~al.}(2015)Radford, Metz, and
  Chintala]{radford2015unsupervised}
Radford, A., Metz, L., and Chintala, S. (2015).
\newblock Unsupervised representation learning with deep convolutional
  generative adversarial networks.
\newblock {\em arXiv preprint arXiv:1511.06434\/}.

\bibitem[Saxe {\em et~al.}(2011)Saxe, Koh, Chen, Bhand, Suresh, and
  Ng]{ICML2011Saxe_551}
Saxe, A., Koh, P.~W., Chen, Z., Bhand, M., Suresh, B., and Ng, A. (2011).
\newblock On random weights and unsupervised feature learning.
\newblock In L.~Getoor and T.~Scheffer, editors, {\em Proceedings of the 28th
  International Conference on Machine Learning (ICML-11)\/}, ICML '11, pages
  1089--1096, New York, NY, USA. ACM.

\bibitem[Visin {\em et~al.}(2015)Visin, Kastner, Courville, Bengio, Matteucci,
  and Cho]{visin15}
Visin, F., Kastner, K., Courville, A.~C., Bengio, Y., Matteucci, M., and Cho,
  K. (2015).
\newblock Reseg: {A} recurrent neural network for object segmentation.

\bibitem[Yu and Koltun(2015)Yu and Koltun]{yu2015multi}
Yu, F. and Koltun, V. (2015).
\newblock Multi-scale context aggregation by dilated convolutions.
\newblock {\em arXiv preprint arXiv:1511.07122\/}.

\bibitem[Zeiler and Fergus(2014)Zeiler and Fergus]{zeiler2014visualizing}
Zeiler, M.~D. and Fergus, R. (2014).
\newblock Visualizing and understanding convolutional networks.
\newblock In {\em Computer vision--ECCV 2014\/}, pages 818--833. Springer.

\bibitem[Zeiler {\em et~al.}(2011)Zeiler, Taylor, and
  Fergus]{zeiler2011adaptive}
Zeiler, M.~D., Taylor, G.~W., and Fergus, R. (2011).
\newblock Adaptive deconvolutional networks for mid and high level feature
  learning.
\newblock In {\em Computer Vision (ICCV), 2011 IEEE International Conference
  on\/}, pages 2018--2025. IEEE.

\end{thebibliography}
\bibliographystyle{natbib}
\end{document}